\documentclass{article}

     \PassOptionsToPackage{numbers, compress}{natbib}

 \usepackage[preprint]{neurips_2025}

\usepackage[utf8]{inputenc} 
\usepackage[T1]{fontenc}    
\usepackage{hyperref}       
\usepackage{url}            
\usepackage{booktabs}       
\usepackage{amsfonts}       
\usepackage{nicefrac}       
\usepackage{microtype}      
\usepackage{graphicx} 
\usepackage{amsmath}

\usepackage[table]{xcolor} 
\usepackage[capitalize]{cleveref}
\usepackage{array}
\usepackage{enumitem}
\usepackage{natbib}
\setcitestyle{numbers,square}
\usepackage{wrapfig}
\usepackage{times}
\usepackage{epsfig}
\usepackage{float}
\usepackage{adjustbox}
\usepackage{caption}
\usepackage{utfsym}
\usepackage{bbding}
\usepackage{makecell}
\usepackage{multirow}
\usepackage{amssymb}
\usepackage{listings}
\newlength\savewidth

\usepackage{subcaption}
\definecolor{backgroundcolor}{RGB}{232, 242, 255}
\definecolor{mygray}{gray}{0.9}

\newcommand{\tablestyle}[2]{\setlength{\tabcolsep}{#1}\renewcommand{\arraystretch}{#2}\centering\footnotesize}

\title{\includegraphics[width=0.09\textwidth]{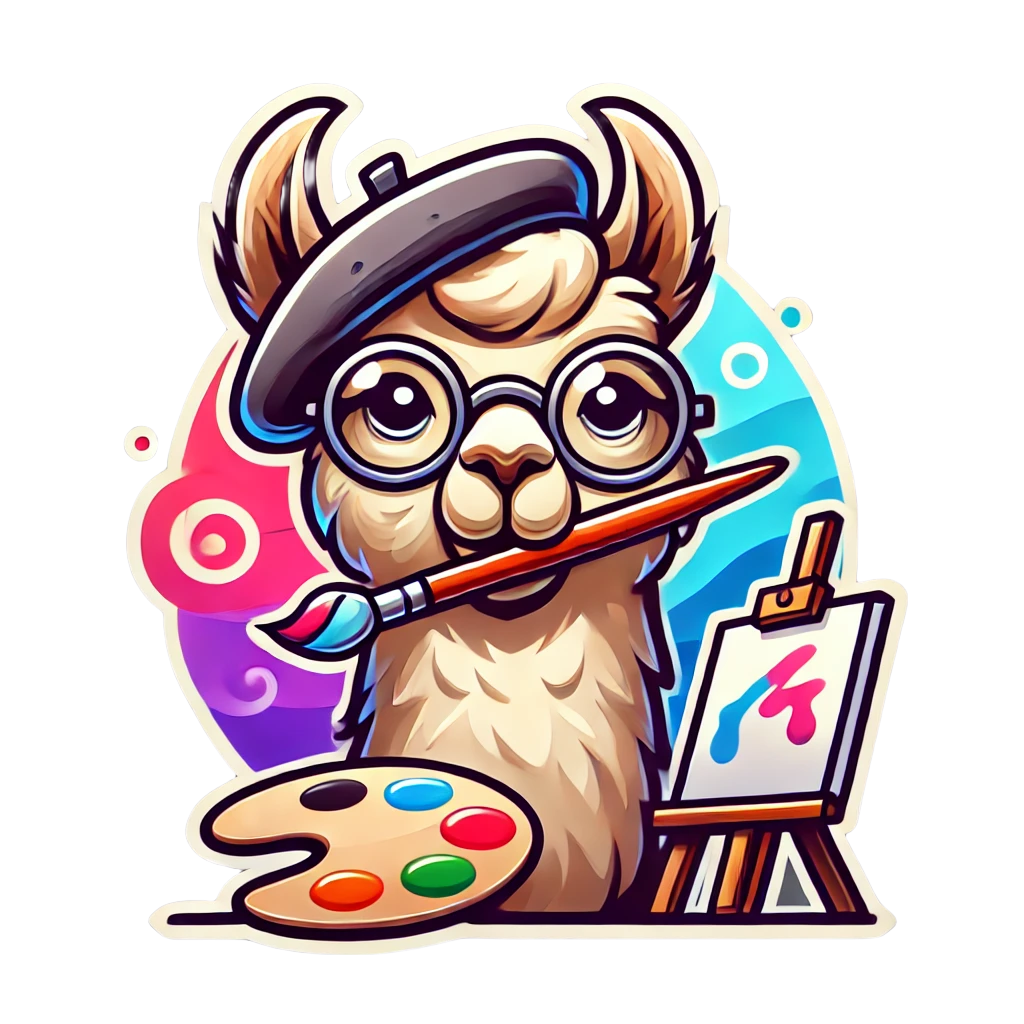} FOCUS: Unified Vision-Language Modeling for Interactive Editing Driven by Referential Segmentation}

\author{
\textbf{Fan Yang}\textsuperscript{\rm 1,2,3}
\textbf{Yousong Zhu}\textsuperscript{\rm 1$\dagger$},
\textbf{Xin Li}\textsuperscript{\rm 3},
\textbf{Yufei Zhan}\textsuperscript{\rm 1,2},
\textbf{Hongyin Zhao}\textsuperscript{\rm 1},\\
\textbf{Shurong Zheng}\textsuperscript{\rm 1,2,3},
\textbf{Yaowei Wang}\textsuperscript{\rm 3},
\textbf{Ming Tang}\textsuperscript{\rm 1,2},
\textbf{Jinqiao Wang}\textsuperscript{\rm 1,2,3,4},
\\
\\
\textsuperscript{\rm 1}Foundation Model Research Center, Institute of Automation, \\
Chinese Academy of Sciences, 
Haidian District, Beijing, China \\
\textsuperscript{\rm 2}School of Artificial Intelligence, \\
University of Chinese Academy of Science, Beijing, China
\\
\textsuperscript{\rm 3}Peng Cheng Laboratory, Shenzhen, China \\
\textsuperscript{\rm 4}Wuhan AI Research, Wuhan, China
\\
}

\begin{document}

\maketitle

\renewcommand{\thefootnote}{\fnsymbol{footnote}}
\footnotetext[2]{is the corresponding authors}

\begin{abstract}

Recent Large Vision Language Models (LVLMs) demonstrate promising capabilities in unifying visual understanding and generative modeling, enabling both accurate content understanding and flexible editing. However, current approaches treat \textbf{\textit{"what to see"}} and \textbf{\textit{"how to edit"}} separately: they either perform isolated object segmentation or utilize segmentation masks merely as conditional prompts for local edit generation tasks, often relying on multiple disjointed models. To bridge these gaps, we introduce FOCUS, a unified LVLM that integrates segmentation-aware perception and controllable object-centric generation within an end-to-end framework. FOCUS employs a dual-branch visual encoder to simultaneously capture global semantic context and fine-grained spatial details. In addition, we leverage a MoVQGAN-based visual tokenizer to produce discrete visual tokens that enhance generation quality. To enable accurate and controllable image editing, we propose a progressive multi-stage training pipeline, where segmentation masks are jointly optimized and used as spatial condition prompts to guide the diffusion decoder. This strategy aligns visual encoding, segmentation, and generation modules, effectively bridging segmentation-aware perception with fine-grained visual synthesis.
Extensive experiments across three core tasks, including multimodal understanding, referring segmentation accuracy, and controllable image generation, demonstrate that FOCUS achieves strong performance by jointly optimizing visual perception and generative capabilities.

\end{abstract}
\section{Introduction}
\label{sec:intro}

Large Vision-Language Models (LVLMs) are becoming a transformative paradigm in artificial intelligence~\cite{chatgpt, claude, claude3, LLaVA}. Through large-scale pretraining, they unify visual and textual modalities and have achieved remarkable progress across a variety of tasks. These models can jointly process images, videos, and natural language within a single architecture, demonstrating strong performance in visual question answering, image captioning, referring segmentation, and conditional generation ~\cite{video-llava, Ferret, pixellm, LLaVA, CogVLM}. The effectiveness of LVLMs largely stems from their ability to align multimodal semantics and generalize across diverse tasks with minimal task-specific adaptation.

\begin{figure}
\centering
\includegraphics[width=1\textwidth]{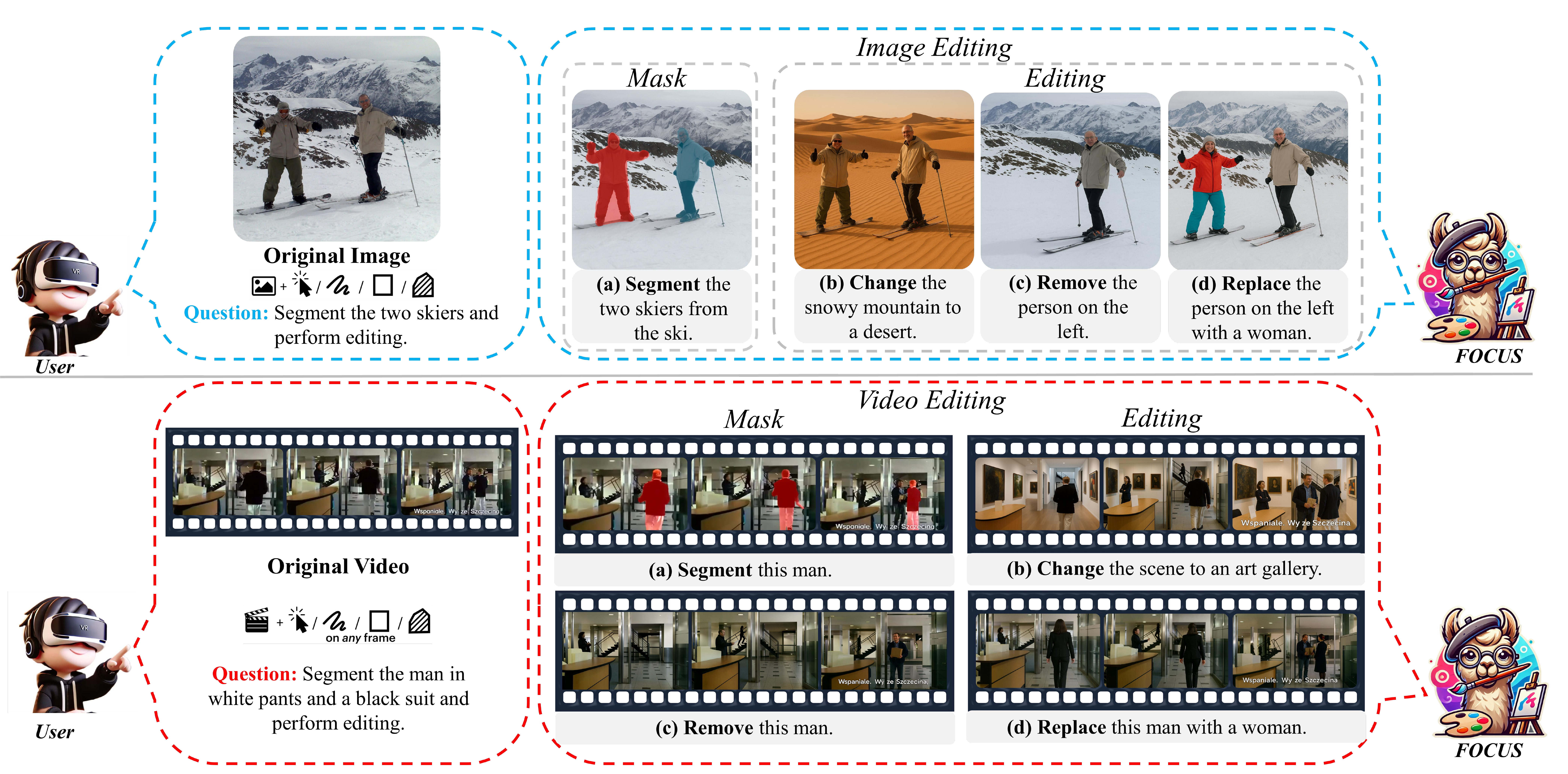}
\caption{\textbf{FOCUS} enables fine-grained segmentation and editing for both \textbf{images} and \textbf{videos} through multimodal user instructions. It supports various region specification formats including \textbf{clicks}, \textbf{scribbles}, \textbf{boxes}, and \textbf{masks}, and for videos, \textbf{annotations on any single frame} suffice to guide full-clip editing. FOCUS can perform detailed region segmentation (e.g., identifying individual people), and supports diverse editing operations such as \textbf{removal, replacement}, and \textbf{scene transformation} across spatial and temporal domains.}
\label{fig:first}
\vspace{-3.5mm}
\end{figure}

\begin{figure}
\centering
\includegraphics[width=1\textwidth]{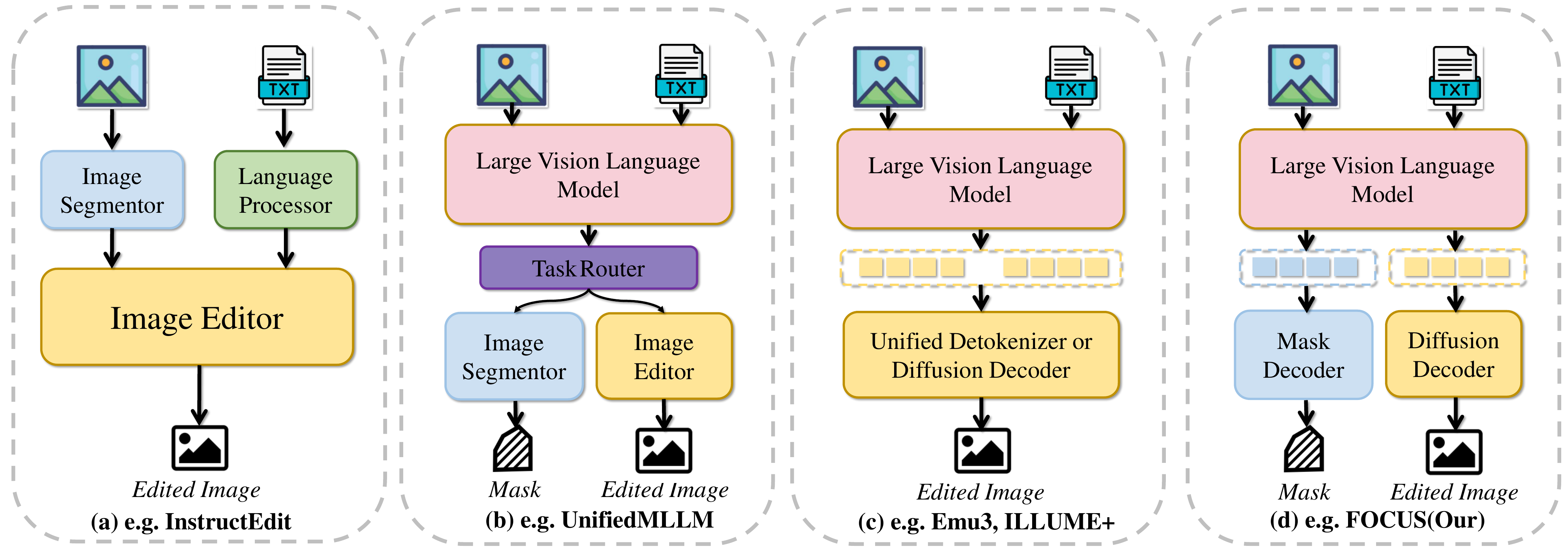}
\caption{Comparison of controllable image editing paradigms. (a) Modular methods rely on separately trained components. (b) Task-routing LLMs orchestrate existing tools without joint modeling. (c) Unified models combine perception and generation but lack fine-grained control. (d) Our FOCUS jointly models segmentation and generation for precise, region-level editing.
}
\label{fig:diff}
\vspace{-3.5mm}
\end{figure}

At the same time, image and video generation technologies~\cite{podell2023sdxlimprovinglatentdiffusion, chen2023pixartalphafasttrainingdiffusion, peebles2023scalablediffusionmodelstransformers, zheng2024opensorademocratizingefficientvideo} have advanced rapidly, enabling the synthesis of high-quality visual content and reshaping how we approach artistic creation and visual communication. These developments have driven increasing demands for controllability and fidelity across domains such as art, industry, and education~\cite{gou2024toratoolintegratedreasoningagent}. To meet these demands, some approaches~\cite{wang2023instructeditimprovingautomaticmasks, couairon2022diffeditdiffusionbasedsemanticimage, chen2023llavainteractiveallinonedemoimage} as is shown in fig.~\ref{fig:diff}(a) attempt to combine image generation models with segmentation decoders and text processing modules through modular design. However, such methods often rely on manually engineered pipelines and lack unified modeling and deep feature-level interaction.
To address these limitations, others method~\cite{li2024unifiedmllmenablingunifiedrepresentation} as in fig.~\ref{fig:diff}(b) leverage the instruction understanding capabilities of LVLMs to dynamically dispatch pretrained expert models through task routing, enabling flexible multi-task adaptation. While these approaches improve tool orchestration, they lack deep cross-modal fusion and joint feature optimization, making it difficult to unify perception and generation effectively. More recent methods\cite{emu, emu2} as is show in fig.~\ref{fig:diff}(c) have begun to construct unified vision-language frameworks that combine image understanding and generation, using the generalization capability of LVLMs to bridge the semantic gap between high-level understanding and low-level synthesis. However, most of these methods still operate at a coarse level of text-driven control and struggle to support fine-grained editing or object-level manipulation.

To overcome these challenges, we propose FOCUS, an end-to-end LVLM framework that unifies segmentation-aware perception and region-controllable generation under natural language guidance, as is shown in fig.~\ref{fig:first}. The core of FOCUS features a dual-branch visual encoder, where a CLIP-like or QwenViT-style encoder extracts global semantic representations, while a hierarchical encoder based on ConvNeXt-L focuses on fine-grained local perception. This structure provides stable multi-scale segmentation support and improves adaptability to varying image resolutions.
To improve visual generation, FOCUS adopts a VQGAN-based visual tokenizer, inspired by ~\cite{wang2024emu3nexttokenpredictionneed, chameleonteam2025chameleonmixedmodalearlyfusionfoundation, huang2025illumeilluminatingunifiedmllm}, which separately models semantic concepts and texture information. To mitigate the inevitable information loss during quantization, we retain the continuous pre-quantization features from the tokenizer as visual inputs to the language model, enabling fine-grained multimodal understanding.
FOCUS is trained through a progressive multi-stage strategy, gradually increasing the input and output resolutions from low to high to ensure stable convergence and final performance. Segmentation masks are jointly optimized and used as spatial condition prompts to guide a diffusion-based generator for pixel-level editing. During both the multimodal pretraining and instruction tuning stages, we introduce diverse and increasingly complex task distributions, with carefully designed instruction formats, to fully enhance the model's perception understanding and generation capabilities.

Experimental results show that FOCUS achieves significant improvements across three core tasks: multimodal understanding, controllable image generation and editing, and referring segmentation. By jointly modeling pixel-level perception and generation, FOCUS demonstrates higher semantic precision and stronger spatial controllability. Further analysis reveals that pixel-level perception plays a crucial role in bridging the gap between high-level understanding and low-level synthesis. These findings confirm the feasibility and effectiveness of unifying segmentation-aware perception and controllable generation within a single LVLM framework, paving the way for interactive, precise, and generalizable multimodal editing systems.


\begin{itemize}
  \item We propose FOCUS, a unified large vision language model that integrates pixel-level perception with region-controllable image generation and editing within an end-to-end framework.

  \item We develop a progressive multi-stage training pipeline that gradually increases image resolution and task complexity. This pipeline enables effective alignment and interaction between the segmentation decoder and the generation module across different scales and modalities.

  \item Extensive experiments on multimodal understanding, referential segmentation, and controllable image editing demonstrate that FOCUS consistently outperforms existing state-of-the-art models in both visual understanding and generation quality.
\end{itemize}


\section{Related work}\label{sec:related}

The landscape of large vision–language models (LVMs) has rapidly evolved, aiming to unify multimodal understanding and generation across images, videos, and texts. A representative example is the Emu series, which introduces autoregressive modeling to predict the next visual or textual token, thus enabling a generalist interface for diverse multimodal tasks such as image captioning, video event understanding, and cross-modal generation. This contrasts with earlier LVM designs (e.g., BLIP~\cite{BLIP-2}, Flamingo~\cite{flamingo}) that bridge frozen vision and language models using separate modality connectors, often focusing solely on text prediction and neglecting direct supervision over visual signals. Emu2~\cite{emu2} further extends this paradigm by positioning itself as an in-context learner, exploiting interleaved video–text data to deliver fine-grained temporal and causal reasoning over multimodal inputs.


While many of these works achieve remarkable integration of multimodal comprehension and generation, they often fall short in interactive and controllable editing. For instance, the LLaVA-Interactive~\cite{llavainteractive} system explores multi-turn, multimodal interactions combining visual chat, segmentation, and editing but largely depends on the synergy of pretrained components without true end-to-end learning. More critically, recent editing frameworks such as InstructEdit emphasize the importance of high-quality segmentation masks by leveraging external Grounded-SAM masks to guide diffusion-based image editing. However, these approaches are not fully end-to-end; they separate mask generation from the editing pipeline, highlighting a major gap in unifying segmentation awareness with controllable generation.

In parallel, dataset-centric efforts like Localized Narratives~\cite{ponttuset2020connectingvisionlanguagelocalized} and Video Localized Narratives~\cite{voigtlaender2023connectingvisionlanguagevideo} have emerged, providing rich multimodal annotations by aligning every word or phrase with specific image or video regions. These resources enable granular referential grounding and enhance the evaluation and training of models that need to capture both spatial and temporal semantics across modalities. Additionally, Describe Anything~\cite{lian2025anythingdetailedlocalizedimage} offers detailed localized captioning, helping bridge the gap between referential understanding and region-specific generation in both image and video domains.

Collectively, these prior works lay the foundation for advancing multimodal understanding, generation, and editing. Yet, there remains an unmet need for a unified architecture that couples segmentation-aware decoding with mask-driven diffusion editing in a fully end-to-end manner. This motivates the development of the proposed framework, which seeks to integrate referential localization, structured mask generation, and controllable editing into a seamless multimodal pipeline.

\section{Methodology}

\begin{figure}
    \centering
    \includegraphics[width=1\textwidth]{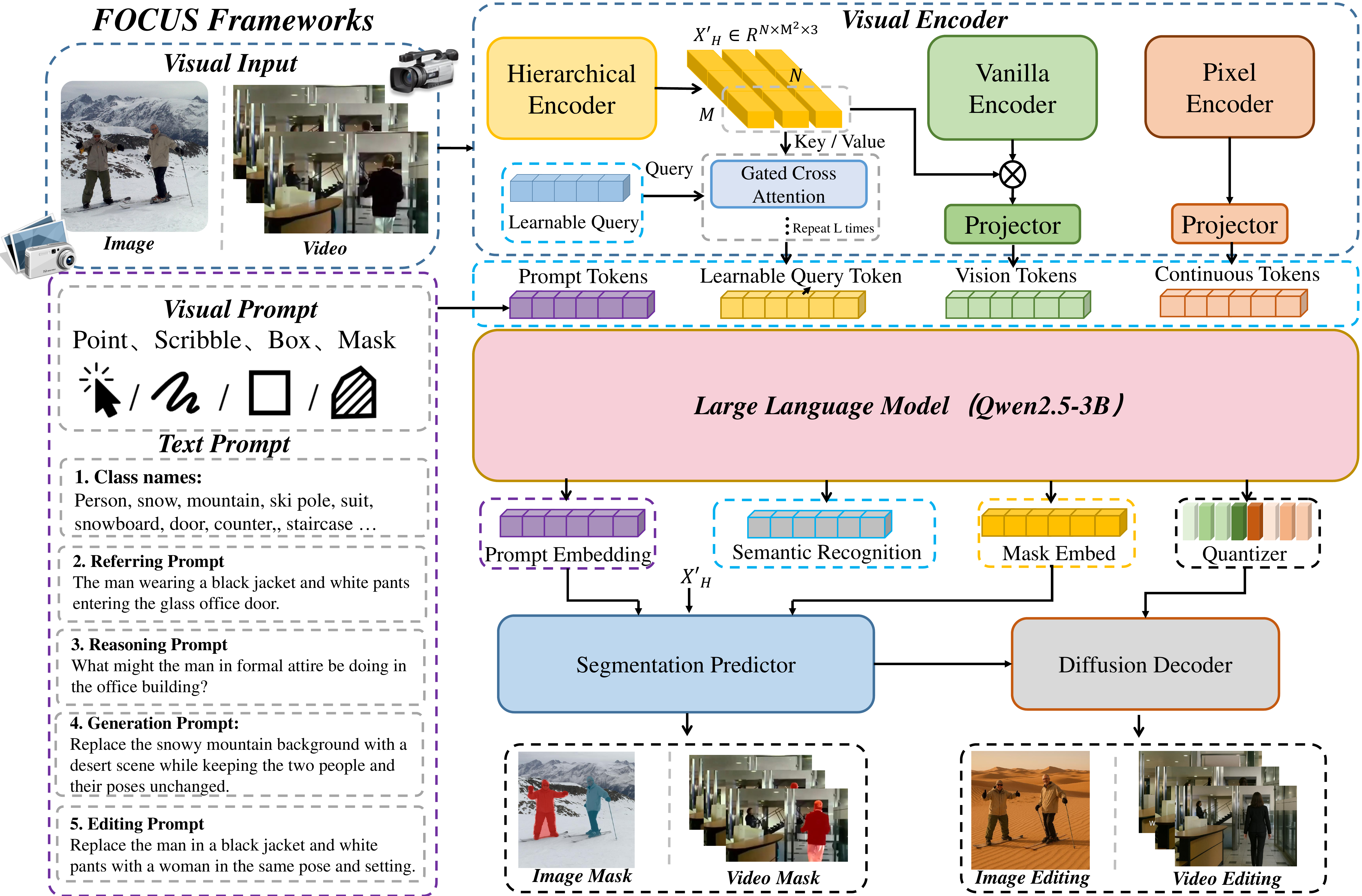}
    \caption{Overview of the FOCUS framework and training pipeline. The left part shows the unified architecture, which integrates a vanilla encoder, a hierarchical encoder, a pixel encoder, a large language model, a segmentation predictor, and a diffusion decoder for fine-grained perception and controllable image generation. }
    \label{fig:main}
    \vspace{-3.5mm}
\end{figure}


As illustrated in Fig.\ref{fig:main}, FOCUS is a unified large vision-language model that integrates pixel-level perception and controllable image generation in an end-to-end framework. The model comprises four core components, covering the entire process from visual encoding to image synthesis. ~\cref{sec:dual_visual_encoders_and_generative_tokenizer} introduce a dual-branch visual encoder and a generation-oriented visual tokenizer are employed to extract global semantic features and multi-scale fine-grained representations from inputs at different resolutions, discretizing the visual content into high-level tokens suitable for downstream editing tasks. Next, in the ~\cref{sec:large_language_model_and_input_schema} a large language model based on Qwen2.5 is introduced to unify multimodal input formats through task prompts, enabling support for a wide range of vision-language tasks. Building on this, in the ~\cref{sec:Segmentation} a segmentation decoder is designed to perform high-precision fine-grained object segmentation. A diffusion-based image generator then synthesizes high-fidelity images guided by spatial segmentation masks and conditioned on discrete visual tokens in the ~\cref{sec:Diffusion}. In the ~\cref{sec:Training} further details the progressive training strategy and the construction of multi-source datasets.


\subsection{Dual Visual Encoders and Generative Tokenizer}
\label{sec:dual_visual_encoders_and_generative_tokenizer}


A fundamental challenge in building unified large vision-language models for both image understanding and generation lies in the significant discrepancy between high-level semantic understanding and low-level fine-grained image synthesis. This discrepancy often leads to mutual interference and optimization conflicts during training. To address this issue, we propose a novel architecture that combines a dual-branch visual encoder with a generative visual tokenizer. This design preserves the ability to model global high-level semantic features while enhancing the representation of fine-grained, multi-scale low-level visual details. Meanwhile, the visual tokenizer discretizes continuous visual information, effectively extending the representational capacity of the image generation module.

Specifically, we process each image at two resolutions. A low-resolution image $X_L \in \mathbb{R}^{3 \times 256 \times 256}$ is fed into a \textbf{Vanilla Encoder} (based on QwenViT~\cite{Qwen-vl}) to capture global semantic information. Simultaneously, a high-resolution image $X_H \in \mathbb{R}^{3 \times 768 \times 768}$ is encoded by a \textbf{Hierarchical Encoder} (ConvNeXt-L~\cite{liu2022convnet}) to obtain high-resolution, multi-scale visual features. These two branches are fused using a cross-attention mechanism to enhance low-resolution features with local detail:
\begin{equation}
X_H' = \text{ConvNeXt}(X_H), \quad X_L' = \text{QwenViT}(X_L)
\end{equation}
\begin{equation}
E'_{\text{img}} = \text{CrossAttn}(X_L', X_H'), \quad
E_{\text{img}} = \text{MLP}(E'_{\text{img}}) + E'_{\text{img}}
\end{equation}

To further inject fine-grained visual information into the language model, we introduce a gated cross-attention adapter that enhances learnable queries with high-resolution visual features at multiple scales. For the $l$-th layer query $h^{(l)}$, and $j$-th scale feature $f_{\text{img}}^{(j)}$, the fusion is formulated as:
\begin{equation}
h^{(l)'} = h^{(l)} + \tanh(\gamma^{(l)}) \cdot \text{CrossAttn}(h^{(l)}, G_p(f_{\text{img}}^{(j)}))
\end{equation}
\begin{equation}
h^{(l)}_{\text{Adapter}} = h^{(l)'} + \tanh(\beta^{(l)}) \cdot \text{FFN}(h^{(l)'})
\end{equation}
where $G_p(\cdot)$ is a projection function aligning the visual feature space, and $\gamma^{(l)}, \beta^{(l)}$ are learnable scaling parameters initialized to zero. This mechanism injects spatial detail while maintaining the efficiency of learnable queries, enhancing both segmentation and generative capabilities.

To enhance the upper bound of image generation quality, we introduce a \textbf{generative visual tokenizer} based on the MoVQGAN~\cite{zheng2022movqmodulatingquantizedvectors} architecture to discretize visual information. The visual tokenizer serves as a key component for unified autoregressive image generation, but its quantization process inevitably introduces information loss. To address this, we use the continuous features before quantization as the visual input to the large language model (LLM), which provides a more informative representation for fine-grained multimodal understanding. The LLM then outputs discrete visual tokens that guide the diffusion model for high-quality image synthesis.

Specifically, we use the fused dual-branch features $E'_{\text{img}}$ as the global semantic representation for generation, which are quantized into discrete semantic tokens and decoded through a lightweight decoder. During the visual tokenizer’s pretraining stage, this semantic branch is supervised with a cosine similarity loss against the original encoder features. A downsampling rate of $28\times$ is adopted to align with mainstream vision-language models and focus on high-level semantic concepts.

We further extend the generative visual tokenizer with a \textbf{pixel branch}, following the standard MoVQGAN design. This branch uses a $16\times$ downsampling rate to preserve fine textures. After quantization, the semantic and pixel tokens are concatenated along the channel dimension and passed to the decoder for image reconstruction. The pixel branch is trained with a combination of L1 loss, perceptual loss, and adversarial loss. We use large codebooks for both branches, with a size of 32{,}768 for the semantic branch and 98{,}304 for the pixel branch. 

Together, the generative visual tokenizer and dual-branch visual encoder construct a unified and expressive visual representation that enables \textbf{FOCUS} to support both fine-grained perception and controllable image generation within a single framework.

\subsection{Large Vision-Language Model and Input Schema}
\label{sec:large_language_model_and_input_schema}

In \textbf{FOCUS}, we adopt \textbf{Qwen2.5-3B}~\cite{yang2024qwen2} as the large language model (LLM) to model unified multimodal inputs from the visual encoder and task instructions. The model takes as input a four-tuple \((E_{\text{img}}, X_S, X_T, h^{(l)}_{\text{Adapter}})\), where \(E_{\text{img}}\) denotes the fused visual features from the dual-branch encoder, \(X_S\) represents the structured textual prompts, \(X_T\) refers to the continuous pixel-level visual features from the pixel encoder, and \(h^{(l)}_{\text{Adapter}}\) is the learnable query feature obtained through multi-scale cross-modal adaptation.

These inputs are fed into the LLM to produce the output representation:
\[
E_O = F_{\text{LLM}}(E_{\text{img}}, X_S, X_T, h^{(l)}_{\text{Adapter}})
\]

From the LLM output \(E_O\), we extract two key components. First, we introduce a decoding constraint that enforces the model to predict the names of all present objects before generating the mask tokens. This prior step encourages the generation of semantically enriched mask tokens, which incorporate global semantic context from the image and are subsequently used by the segmentation module to produce masks.

In addition, the LLM supports autoregressive prediction of discrete visual tokens. Following a semantic-first strategy, the model first generates semantic tokens to define global content structure, then generates pixel tokens to recover fine visual textures. This two-stage decoding improves the alignment between textual instructions and generated visual content.

\paragraph{Prompt Design.}

The input prompt schema consists of two components: the \textit{task instruction prompt} \(S_I\) and the \textit{condition prompt} \(S_C\). The task instruction prompt specifies the objective of the model in natural language, guiding the LLM to perform segmentation, generation, or editing accordingly. For example, in class-based segmentation tasks such as panoptic, open-vocabulary, or video instance segmentation, the instruction can be \textit{“Please segment all the positive objects by the following candidate categories.”} In referring or reasoning segmentation tasks (e.g., RES~\cite{refclef, refcoco, refcocog, grefcoco}, R-VOS~\cite{li2023robustreferringvideoobject}, ReasonVOS~\cite{yan2024visareasoningvideoobject}), the instruction becomes \textit{“Please segment the target referred to by the language description.”} For visual-guided tasks such as interactive or video object segmentation, the instruction is phrased as “Please segment according to the given visual reference regions.”

The condition prompt provides additional information specific to each task. In class-based segmentation, it lists the candidate category labels; in referring segmentation, it provides a natural language expression; in visual-guided settings, it includes pooled region features sampled from CLIP embeddings at specified coordinates. For image generation tasks, the condition prompt consists of a text description such as \textit{“a cat sitting on a couch,”} while for image editing tasks, it contains language-based or spatial referring expressions like \textit{“replace the person on the right with a dog.”} The condition prompt not only supplies contextual information but also serves as an implicit classifier for category-aware mask prediction. 

\subsection{Segmentation Module}
\label{sec:Segmentation}

The segmentation predictor \(F_p\) takes three types of inputs: task-specific prompt embeddings \(\{E_P^k\}_{k=1}^{K}\), semantic-enhanced mask token embeddings \(\{E_Q^j\}_{j=1}^{N}\) from the LLM output, and multi-scale visual features \(f_{\text{img}} = X_H'\) from the visual encoder. Here, \(K\) denotes the number of candidate categories, and \(N\) represents the number of mask proposals. These inputs are fused to predict segmentation outputs in the following form:
\begin{equation}
\{(m_j, z_j, e_j)\}_{j=1}^{N} = F_p\left(\{E_P^k\}_{k=1}^{K}, \{E_Q^j\}_{j=1}^{N}, f_{\text{img}}\right),
\end{equation}
where \(m_j \in \mathbb{R}^{H \times W}\) denotes the \(j\)-th predicted binary mask, \(z_j \in \mathbb{R}^{K}\) is the associated category score vector, and \(e_j \in \mathbb{R}^{D}\) is an optional instance-level embedding produced by an auxiliary embedding head to enable temporal association in video segmentation tasks.

For video tasks, we adopt a frame-by-frame processing strategy to generate frame-level segmentation results, ensuring efficient training and inference while maintaining temporal consistency.

To support multi-scale supervision in the subsequent image generation stage, the segmentation module consistently produces fixed-resolution masks based on \(X_H'\), providing stable spatial guidance. These masks are then rescaled to match the resolution used by the diffusion model, enabling effective alignment between segmentation outputs and the image synthesis pipeline.

\subsection{Diffusion Decoder for Controllable Generation}
\label{sec:Diffusion}
To enable high-quality and region-controllable image generation, \textsc{FOCUS} incorporates a latent diffusion decoder initialized from SDXL. The generation process is formulated as denoising over latent variables, conditioned on structured visual prompts. We utilize semantic tokens \(z_{\text{sem}}\) and pixel-level tokens \(z_{\text{pix}}\) generated by the large language model. These tokens are mapped into continuous embeddings via a learned codebook and concatenated with noisy latents before being passed into a UNet-based denoising backbone.

To achieve spatial control, we leverage predicted segmentation masks \(m_j \in \mathbb{R}^{H \times W}\) from the segmentation module. These masks are first downsampled to match the latent spatial resolution, yielding \(\tilde{m}_j \in \mathbb{R}^{H' \times W'}\), where \(H' \times W'\) denotes the latent feature resolution. The downsampled mask is then flattened and passed through a linear projection layer to obtain a spatial guidance sequence \(f_m \in \mathbb{R}^{L \times C}\), where \(L = H' \times W'\) and \(C\) is the channel dimension. This sequence is injected into the UNet through cross-attention at intermediate layers to guide regional generation. The interaction is formalized as:

\begin{equation}
\hat{z}_t = \text{CrossAttn}(\phi(z_t), f_m), \quad z_{t+1} = \text{UNet}(\hat{z}_t),
\end{equation}

where \(\phi(z_t)\) denotes the projected latent features at denoising timestep \(t\), and \(f_m\) provides the spatial condition derived from \(\tilde{m}_j\).

We also experimented with using mask token embeddings \(\{E_Q^j\}\) as alternative conditioning inputs, but found that direct spatial control through explicit masks yields better localization and more faithful region editing in our experiments.

\subsection{Training Procedure and Data Composition}
\label{sec:Training}

To support unified pixel-level understanding, high-level multimodal reasoning, and spatially controllable image generation, \textsc{FOCUS} adopts a progressive four-stage training paradigm that incrementally builds visual tokenization, vision-language alignment, and conditional generation capabilities.

\vspace{0.5em}
\noindent \textbf{Stage 1: Pretraining of Dual-Branch Visual Tokenizer and Diffusion Decoder.}  
We first train a semantic branch and a pixel branch to discretize input images into semantic and texture tokens using SimVQ quantizers. A progressive resolution strategy is adopted, starting from $256 \times 256$ and gradually increasing to $512 \times 512$. A large-scale image-text dataset of approximately 58M pairs is constructed based on resolution constraints and visual diversity. On top of this, we pretrain a latent diffusion decoder using a 10M image subset, enabling high-fidelity reconstruction from discrete tokens. This stage focuses purely on image modeling and does not involve segmentation or controllable generation.

\vspace{0.5em}
\noindent \textbf{Stage 2: Visual-Language Adapter Warmup.}  
To align visual features with the input space of the large language model (LLM), we train projection heads for both the vanilla encoder and the pixel encoder, along with learnable queries and gated cross-attention modules in the hierarchical encoder branch. All visual backbone encoders are kept frozen. Training is conducted at $256 \times 256$ resolution with standard language modeling loss.

\vspace{0.5em}
\noindent \textbf{Stage 3: Multimodal Pretraining with Segmentation-Aware Alignment.}  
In this stage, we jointly train the LLM, visual adapters, and the mask decoder to improve the model's perception of segmentation structures and enhance multimodal generation capability. Training is performed in two phases with input resolutions of $256 \times 256$ and $512 \times 512$. Supervision includes token-level cross-entropy loss, segmentation mask supervision (Dice + BCE), and image reconstruction loss (L2) from diffusion outputs. The training corpus includes multimodal data spanning image-text pairs, segmentation tasks, and vision-language reasoning. Full dataset details are presented in appendix.

\vspace{0.5em}
\noindent \textbf{Stage 4: Instruction Tuning for Region-Controlled Editing.}  
This stage introduces region-level controllability for editing and generation tasks. We jointly train the LLM, the mask decoder, and cross-attention layers within the diffusion model, while freezing the visual encoders, visual tokenizer, and VAE components. Input images are resized using a bucketed aspect ratio cropping strategy with total pixel counts ranging from $512^2$ to $1024^2$. Supervision includes token prediction (cross-entropy), segmentation mask prediction (Dice + BCE), and image reconstruction (L2 loss). Detailed task and dataset configurations are summarized in appendix.

\section{Experiments}
\label{sec:experiments}


We conduct comprehensive experiments to evaluate the performance of FOCUS across three representative tasks: multimodal understanding, referential segmentation, and generation and controllable editing. All experiments are designed to validate the model’s ability to unify perception and generation in an end-to-end framework, with strong alignment to natural language instructions, fine-grained localization, and object-aware editing fidelity.

\subsection{Implementation Details}
\label{sec:exp_setup}

In our experiments, we adopt Qwen2.5 as the backbone large language model (LLM). The visual encoder consists of two branches: the semantic decoder is composed of four attention blocks with 2D relative positional encoding (2D-RoPE), while the pixel encoder and decoder follow a MoVQGAN-based architecture with base channel dimensions of 128 and 384, respectively. The codebook size is set to 32,768 for the semantic branch and 98,304 for the pixel branch, with both using a code dimension of 32. We employ the AdamW optimizer without weight decay and use a constant learning rate across the visual encoder, diffusion decoder, and the large vision language model. Detailed training hyperparameters for each component are summarized in the supplementary materials. The training of the Dual-Branch Visual Tokenizer and the diffusion decoder each took approximately 3 days on a computing cluster, while the 3B-parameter MLLM required around 13 days to complete the three-stage training process.

\begin{table}[ht]
    \centering
    \caption{FOCUS performance on general and document-oriented benchmarks.}
    \scalebox{0.52}{
    \begin{tabular}{llcccccccccccc}
    \hline
                             & \multicolumn{1}{l|}{}                                   & \multicolumn{7}{c|}{General}                                                                                                                                                                                                                                                                                                           & \multicolumn{5}{c}{Doc}                                                                                                                                                                                                                  \\
    \multirow{-2}{*}{Method} & \multicolumn{1}{l|}{\multirow{-2}{*}{LLM}}              & \multicolumn{1}{l}{POPE}                     & \multicolumn{1}{l}{MMBench}                  & \multicolumn{1}{l}{SEED}                     & \multicolumn{1}{l}{MME-P}                    & \multicolumn{1}{l}{MM-Vet}                   & \multicolumn{1}{l}{MMMU}                     & \multicolumn{1}{l|}{AI2D}                    & \multicolumn{1}{l}{VQA-text}                 & \multicolumn{1}{l}{ChartQA}                  & \multicolumn{1}{l}{DocVQA}                   & \multicolumn{1}{l}{InfoVQA}                  & \multicolumn{1}{l}{OCRBench}                 \\ \hline
    \multicolumn{14}{c}{Understanding Only}                                                                                                                                                                                                                                                                                                                                                                                                                                                                                                                                                                                                                                \\ \hline
    InstructBLIP~\cite{Instructblip}& \multicolumn{1}{l|}{Vicuna-7B~\cite{video-llava}}                          & -                                            & 36.0                                         & 53.4                                         & -                                            & 26.2                                         & 30.6                                         & \multicolumn{1}{c|}{33.8}                    & 50.1                                         & 12.5                                         & 13.9                                         & -                                            & 276                                          \\
    Qwen-VL-Chat~\cite{Qwen-vl}             & \multicolumn{1}{l|}{Qwen-7B}                            & -                                            & 60.6                                         & 58.2                                         & 1487.5                                       & -                                            & 35.9                                         & \multicolumn{1}{c|}{45.9}                    & 61.5                                         & 66.3                                         & 62.6                                         & -                                            & 488                                          \\
    LLaVA-1.5~\cite{LLaVA}                & \multicolumn{1}{l|}{Vicuna-7B}                          & 85.9                                         & 64.3                                         & 58.6                                         & 1510.7                                       & 31.1                                         & 35.4                                         & \multicolumn{1}{c|}{54.8}                    & 58.2                                         & 18.2                                         & 28.1                                         & 25.8                                         & 318                                          \\
    ShareGPT4V~\cite{chen2023sharegpt4vimprovinglargemultimodal}               & \multicolumn{1}{l|}{Vicuna-7B}                          & -                                            & 68.8                                         & 69.7                                         & 1567.4                                       & 37.6                                         & 37.2                                         & \multicolumn{1}{c|}{58}                      & 60.4                                         & 21.3                                         & -                                            & -                                            & 371                                          \\
    LLaVA-NeXT~\cite{li2024llavaonevisioneasyvisualtask}               & \multicolumn{1}{l|}{Vicuna-7B}                          & 86.5                                         & 67.4                                         & 64.7                                         & 1519                                         & 43.9                                         & 35.1                                         & \multicolumn{1}{c|}{66.6}                    & 64.9                                         & 54.8                                         & 74.4                                         & 37.1                                         & 532                                          \\
    Emu3-Chat~\cite{wang2024emu3nexttokenpredictionneed}                & \multicolumn{1}{l|}{8B from scratch}                    & 85.2                                         & 58.2                                         & 68.2                                         & -                                            & 37.2                                         & 31.6                                         & \multicolumn{1}{c|}{70.0}                    & 64.7                                         & 68.6                                         & 76.3                                         & 43.8                                         & 687                                          \\ \hline
    \multicolumn{14}{c}{Unify Understanding and Generation}                                                                                                                                                                                                                                                                                                                                                                                                                                                                                                                                                                                                                \\ \hline
    Unified-IO~\cite{lu2022unifiediounifiedmodelvision}               & \multicolumn{1}{l|}{6.8B from scratch}                  & 87.7                                         & -                                            & 61.8                                         & -                                            & -                                            & -                                            & \multicolumn{1}{c|}{-}                       & -                                            & -                                            & -                                            & -                                            & -                                            \\
    Chameleon~\cite{chameleonteam2025chameleonmixedmodalearlyfusionfoundation}                & \multicolumn{1}{l|}{7B from scratch}                    & -                                            & -                                            & -                                            & -                                            & 8.3                                          & 22.4                                         & \multicolumn{1}{c|}{-}                       & -                                            & -                                            & -                                            & -                                            & -                                            \\
    LWM~\cite{li2023robustreferringvideoobject}                      & \multicolumn{1}{l|}{LLaVA-2-7B}                         & 75.2                                         & -                                            & -                                            & -                                            & 9.6                                          & -                                            & \multicolumn{1}{c|}{-}                       & 18.8                                         & -                                            & -                                            & -                                            & -                                            \\
    Show-o~\cite{xie2024showosingletransformerunify}                   & \multicolumn{1}{l|}{Phi-1.5B}                           & 73.8                                         & -                                            & -                                            & 948.4                                        & -                                            & 25.1                                         & \multicolumn{1}{c|}{-}                       & -                                            & -                                            & -                                            & -                                            & -                                            \\
    VILA-U (256) ~\cite{wu2025vilauunifiedfoundationmodel}            & \multicolumn{1}{l|}{LLaMA-2-7B}                         & 83.9                                         & -                                            & 56.3                                         & 1336.2                                       & 27.7                                         & -                                            & \multicolumn{1}{c|}{-}                       & 48.3                                         & -                                            & -                                            & -                                            & -                                            \\
    VILA-U (384)   ~\cite{wu2025vilauunifiedfoundationmodel}          & \multicolumn{1}{l|}{LLaMA-2-7B}                         & 85.8                                         & -                                            & 59                                           & 1401.8                                       & 33.5                                         & -                                            & \multicolumn{1}{c|}{-}                       & 60.8                                         & -                                            & -                                            & -                                            & -                                            \\
    Janus~\cite{wu2024janusdecouplingvisualencoding}                    & \multicolumn{1}{l|}{DeepSeek-LLM-1.3B}                  & 87.0                                         & 69.4                                         & 63.7                                         & 1338.0                                       & 34.3                                         & 30.5                                         & \multicolumn{1}{c|}{-}                       & -                                            & -                                            & -                                            & -                                            & -                                            \\
    Janus-Pro-1B~\cite{chen2025janusprounifiedmultimodalunderstanding}             & \multicolumn{1}{l|}{DeepSeek-LLM-1.3B}                  & 86.2                                         & 75.5                                         & 68.3                                         & 1444.0                                       & 39.8                                         & 36.3                                         & \multicolumn{1}{c|}{-}                       & -                                            & -                                            & -                                            & -                                            & -                                            \\
    Janus-Pro-7B~\cite{chen2025janusprounifiedmultimodalunderstanding}             & \multicolumn{1}{l|}{DeepSeek-LLM-7B}                    & 87.4                                         & 79.2                                         & 72.1                                         & 1567.1                                       & 50.0                                         & 41.0                                         & \multicolumn{1}{c|}{-}                       & -                                            & -                                            & -                                            & -                                            & -                                            \\
    ILLUME+~\cite{huang2025illumeilluminatingunifiedmllm}                  & \multicolumn{1}{l|}{Qwen2.5-3B}                         & 87.6                                         & 80.8                                         & 73.3                                         & 1414.0                                       & 40.3                                         & 44.3                                         & \multicolumn{1}{c|}{74.2}                    & 69.9                                         & 69.9                                         & 80.8                                         & 44.1                                         & 672                                          \\ \hline
    \rowcolor[HTML]{CBCEFB} 
    FOCUS & \multicolumn{1}{l|}{\cellcolor[HTML]{CBCEFB}Qwen2.5-3B} & \multicolumn{1}{c}{\cellcolor[HTML]{CBCEFB}\textbf{88.0}} & \multicolumn{1}{c}{\cellcolor[HTML]{CBCEFB}\textbf{81.5}} & \multicolumn{1}{c}{\cellcolor[HTML]{CBCEFB}\textbf{73.9}} & \multicolumn{1}{c}{\cellcolor[HTML]{CBCEFB}\textbf{1570.3}} & \multicolumn{1}{c}{\cellcolor[HTML]{CBCEFB}\textbf{50.2}} & \multicolumn{1}{c}{\cellcolor[HTML]{CBCEFB}\textbf{44.9}} & \multicolumn{1}{c}{\cellcolor[HTML]{CBCEFB}\textbf{74.5}} & \multicolumn{1}{c}{\cellcolor[HTML]{CBCEFB}\textbf{70.4}} & \multicolumn{1}{c}{\cellcolor[HTML]{CBCEFB}\textbf{70.3}} & \multicolumn{1}{c}{\cellcolor[HTML]{CBCEFB}\textbf{81.1}} & \multicolumn{1}{c}{\cellcolor[HTML]{CBCEFB}\textbf{44.3}} & \multicolumn{1}{c}{\cellcolor[HTML]{CBCEFB}\underline{678}}  \\ \hline
    \end{tabular}}
    \label{tab:und}
    \end{table}

\subsection{Multimodal understanding}
\label{sec:dialogue_eval}


To evaluate the multimodal understanding capabilities of our model, we conduct systematic evaluations on two categories of widely-used benchmarks, as is show in table ~\ref{tab:und}: (1) General benchmarks, including POPE, MMBench, SEED, MME-P, MM-Vet, MMMU, and AI2D; and (2) Document-oriented benchmarks, including VQA-text, ChartQA, DocVQA, InfoVQA, and OCRBench. Experimental results show that, despite using only a 3B-parameter model, FOCUS achieves performance comparable to state-of-the-art unified models such as Janus-Pro-7B and ILLUME-7B, and notably outperforms ILLUME+ with the same parameter scale. This performance gain is largely attributed to the incorporation of multi-scale high-resolution features and segmentation masks, which significantly enhance pixel-level perception.

\subsection{Controllable Generation and Editing}
\label{sec:gen_eval}

Multimodal image generation. To evaluate the multimodal visual generation capability, we use the MJHQ-30K, GenAI-bench and GenEval benchmarks in the table ~\ref{tab:gen}. For MJHQ-30K, we adopt the Fréchet Inception Distance (FID) metric on 30K generated images compared to 30K high-quality images, measuring the generation quality and diversity. GenAI-bench and GenEval are challenging text-to-image generation benchmarks designed to reflect the consistency between text descriptions and generated images. We compare FOCUS with previous state-ofthe-art multimodal generation-only and unified models. This highlights the superior generation quality and diversity enabled by our diffusion-based approach. Additionally, FOCUS achieves competitive results on the GenAI-bench and GenEval benchmarks and attains the highest accuracy in advanced categories on GenAI-bench, demonstrating its ability to understand and generate images from complex text descriptions. Figure 7 shows more results of FOCUS on generating flexible resolution images.
\begin{table}[ht]
\centering
\caption{
\textbf{Evaluation results on multimodal image generation benchmarks.}
}
    \scalebox{0.62}{
    \tablestyle{7pt}{1}
    \begin{tabular}{lcccccccccccc}
    \hline
                                          &                                    & \multicolumn{1}{c|}{}                                & \multicolumn{1}{c|}{\textit{\textbf{MJHQ30K}}}         & \multicolumn{2}{c|}{\textit{\textbf{GenAI-bench}}}                     & \multicolumn{7}{c}{\textit{\textbf{GenEval}}}                                                                 \\
    \multirow{-2}{*}{\textbf{Method}}     & \multirow{-2}{*}{\textbf{Params.}} & \multicolumn{1}{c|}{\multirow{-2}{*}{\textbf{Type}}} & \multicolumn{1}{c|}{FID}                               & Basic         & \multicolumn{1}{c|}{Adv.}                              & Overall       & Single Obj    & Two Obj.      & Counting      & Colors        & Position      & Color Attri.  \\ \hline
    \multicolumn{13}{c}{\textit{Generation Only}}                                                                                                                                                                                                                                                                                                                                       \\
    SDv1.5~\cite{Rombach_2022_CVPR}         & 0.9B                               & \multicolumn{1}{c|}{Diffusion}                       & \multicolumn{1}{c|}{-}                                 & -             & \multicolumn{1}{c|}{-}                                 & 0.43          & 0.97          & 0.38          & 0.35          & 0.76          & 0.04          & 0.06          \\
    PixArt-$\alpha$~\cite{chen2023pixartalphafasttrainingdiffusion} & 0.6B                               & \multicolumn{1}{c|}{Diffusion}                       & \multicolumn{1}{c|}{6.14}                              & -             & \multicolumn{1}{c|}{-}                                 & 0.48          & 0.98          & 0.45          & 0.44          & 0.08          & 0.07          & 0.07          \\
    SDXL~\cite{podell2023sdxlimprovinglatentdiffusion}            & 2.6B                               & \multicolumn{1}{c|}{Diffusion}                       & \multicolumn{1}{c|}{9.55}                              & \textbf{0.83} & \multicolumn{1}{c|}{0.63}                              & 0.55          & 0.98          & 0.41          & 0.48          & 0.15          & 0.17          & 0.23          \\
    Emu3-Gen~\cite{wang2024emu3nexttokenpredictionneed}            & 8B                                 & \multicolumn{1}{c|}{Autoreg.}                        & \multicolumn{1}{c|}{-}                                 & -             & \multicolumn{1}{c|}{-}                                 & 0.54          & 0.98          & 0.71          & 0.34          & 0.81          & 0.17          & 0.21          \\ \hline
    \multicolumn{13}{c}{\textit{Unify Understanding and Generation}}                                                                                                                                                                                                                                                                                                                    \\
    Chameleon~\cite{chameleonteam2025chameleonmixedmodalearlyfusionfoundation}      & 7B                                 & \multicolumn{1}{c|}{Autoreg.}                        & \multicolumn{1}{c|}{-}                                 & -             & \multicolumn{1}{c|}{-}                                 & 0.39          & -             & -             & -             & -             & -             & -             \\
    LWM~\cite{liu2025worldmodelmillionlengthvideo}                    & 7B                                 & \multicolumn{1}{c|}{Autoreg.}                        & \multicolumn{1}{c|}{17.77}                             & 0.63          & \multicolumn{1}{c|}{0.53}                              & 0.47          & 0.93          & 0.41          & 0.46          & 0.79          & 0.09          & 0.15          \\
    Show-o~\cite{xie2024showosingletransformerunify}             & 1.5B                               & \multicolumn{1}{c|}{Autoreg.}                        & \multicolumn{1}{c|}{15.18}                             & 0.70          & \multicolumn{1}{c|}{0.60}                              & 0.45          & 0.95          & 0.52          & 0.49          & 0.83          & 0.28          & 0.30          \\
    VILA-U(256)~\cite{wu2025vilauunifiedfoundationmodel}           & 7B                                 & \multicolumn{1}{c|}{Autoreg.}                        & \multicolumn{1}{c|}{12.81}                             & 0.72          & \multicolumn{1}{c|}{0.64}                              & -             & -             & -             & -             & -             & -             & -             \\
    VILA-U(384)~\cite{wu2025vilauunifiedfoundationmodel}           & 7B                                 & \multicolumn{1}{c|}{Autoreg.}                        & \multicolumn{1}{c|}{7.69}                              & 0.71          & \multicolumn{1}{c|}{0.66}                              & -             & -             & -             & -             & -             & -             & -             \\
    Janus~\cite{wu2024janusdecouplingvisualencoding}                & 7B                                 & \multicolumn{1}{c|}{Autoreg.}                        & \multicolumn{1}{c|}{10.1}                              & -             & \multicolumn{1}{c|}{-}                                 & -             & -             & -             & -             & -             & -             & -             \\
    Janus-Pro-1B~\cite{chen2025janusprounifiedmultimodalunderstanding}         & 1.3B                               & \multicolumn{1}{c|}{Autoreg.}                        & \multicolumn{1}{c|}{-}                                 & 0.73          & \multicolumn{1}{c|}{0.68}                              & 0.61 & 0.99 & 0.82          & 0.51          & 0.86          & 0.39          & 0.26          \\
    Janus-Pro-7B~\cite{chen2025janusprounifiedmultimodalunderstanding}& 7B& \multicolumn{1}{c|}{Autoreg.}& \multicolumn{1}{c|}{-}& \underline{0.80} & \multicolumn{1}{c|}{0.69}& 0.59& \textbf{0.90} & 0.59& \textbf{0.90}& \textbf{0.79} & 0.66& 0.25          \\
    ILLUME+~\cite{huang2025illumeilluminatingunifiedmllm}& 3B& \multicolumn{1}{c|}{Autoreg.}& \multicolumn{1}{c|}{\underline{6.00}}& 0.72          & \multicolumn{1}{c|}{\underline{0.71}}& \underline{0.72} & \underline{1.00} & \textbf{0.99} & \underline{0.88} & 0.62 & \textbf{0.84} & \underline{0.53} \\ \hline
    \rowcolor[HTML]{CBCEFB} 
    FOCUS&  3B& \multicolumn{1}{c|}{\cellcolor[HTML]{CBCEFB}Autoreg.}& \multicolumn{1}{c|}{\cellcolor[HTML]{CBCEFB}\textbf{6.05}} & {\cellcolor[HTML]{CBCEFB}\textbf{0.83}}& \multicolumn{1}{c|}{\cellcolor[HTML]{CBCEFB}\textbf{0.72}} & \textbf{0.75}     & \textbf{1.20}     & \underline{0.98}     & 0.87    & \underline{0.66}     & \underline{0.81}     & \textbf{0.57}     \\ \hline
    \end{tabular}}
    \label{tab:gen}
\end{table}

Multimodal image editing. To assess the multimodal image editing capability of our method, we evaluate it on the Emu Edit benchmark and report the CLIP-I, CLIP-T, CLIP-DIR and DINO scores. The CLIP-I and DINO scores measure the model’s ability to preserve elements from the source image, while the CLIP-T and CLIP-DIR score measures the consistency between the output image and the target caption. As illustrated in the table ~\ref{tab:editing} , our model demonstrates strong performance in image editing tasks, surpassing specialized models, particularly in the CLIP-T metric. This indicates that the unified model’s superior understanding enhances its ability to interpret editing instructions,
resulting in more precise modifications. 

\begin{table}[ht]
    \centering
    \begin{minipage}{0.45\textwidth}
        \centering
        \caption{\textbf{Comparisons with other referring segmentation.}}
        \LARGE
        \tablestyle{3pt}{1}
        \resizebox{1.0\textwidth}{!}{
        \begin{tabular}{lccccccccccc}
\hline
                         & \multicolumn{3}{c|}{RefCOCO}                                                                            & \multicolumn{3}{c|}{Refcoco+}                                                                           & \multicolumn{2}{c|}{Refcocog}                                               & \multicolumn{3}{c}{gRefCOCO}                                                    \\ \cline{2-12} 
\multirow{-2}{*}{Method} & \multicolumn{1}{l}{Val} & \multicolumn{1}{l}{testA} & \multicolumn{1}{l|}{testB}                        & \multicolumn{1}{l}{Val} & \multicolumn{1}{l}{testA} & \multicolumn{1}{l|}{testB}                        & \multicolumn{1}{l}{Val} & \multicolumn{1}{l|}{Test}                         & \multicolumn{1}{l}{Val} & \multicolumn{1}{l}{testA} & \multicolumn{1}{l}{testB} \\ \hline
\multicolumn{12}{c}{Segmentation Specialist}                                                                                                                                                                                                                                                                                                                                                                 \\ \hline
CRIS~\cite{VCR}                     & 70.5                    & 73.2                      & \multicolumn{1}{c|}{66.1}                         & 62.3                    & 68.1                      & \multicolumn{1}{c|}{53.7}                         & 59.9                    & \multicolumn{1}{c|}{60.4}                         & 55.3                    & 63.8                      & 51.0                      \\
LAVT~\cite{lavt}                     & 72.7                    & 75.8                      & \multicolumn{1}{c|}{68.8}                         & 62.1                    & 68.4                      & \multicolumn{1}{c|}{55.1}                         & 61.2                    & \multicolumn{1}{c|}{62.1}                         & 57.6                    & 65.3                      & 55.0                      \\
PolyFormer-B~\cite{PolyFormer}             & 74.8                    & 76.6                      & \multicolumn{1}{c|}{71.1}                         & 67.6                    & 72.9                      & \multicolumn{1}{c|}{58.3}                         & 67.8                    & \multicolumn{1}{c|}{69.1}                         & -                       & -                         & -                         \\ \hline
\multicolumn{12}{c}{LVLM-based Segmentation Network}                                                                                                                                                                                                                                                                                                                                                         \\ \hline
LISA-7B~\cite{lisa}                  & 74.9                    & 79.1                      & \multicolumn{1}{c|}{72.3}                         & 65.1                    & 70.9                      & \multicolumn{1}{c|}{58.1}                         & 67.9                    & \multicolumn{1}{c|}{70.6}                         & 38.7                    & 52.6                      & 44.8                      \\
PixelLM7B~\cite{pixellm}                & 73.0                    & 76.5                      & \multicolumn{1}{c|}{68.2}                         & 66.3                    & 71.7                      & \multicolumn{1}{c|}{58.3}                         & 69.3                    & \multicolumn{1}{c|}{70.5}                         & -                       & -                         & -                         \\
PSALM~\cite{zhang2024psalmpixelwisesegmentationlarge}                    & 83.6                    & 84.7                      & \multicolumn{1}{c|}{81.6}                         & 72.9                    & 75.5                      & \multicolumn{1}{c|}{70.1}                         & 73.8                    & \multicolumn{1}{c|}{74.4}                         & 42.0                    & 52.4                      & 50.6                      \\
HyperSeg~\cite{wei2024hyperseguniversalvisualsegmentation}& \textbf{84.8}& \underline{85.7}& \multicolumn{1}{c|}{\textbf{83.4}}& \textbf{79.0}& \underline{83.5}& \multicolumn{1}{c|}{\textbf{75.2}}& \textbf{79.4}& \multicolumn{1}{c|}{\underline{78.9}}& 47.5& 57.3& 52.5\\ \hline
\rowcolor[HTML]{CBCEFB} 
FOCUS& \underline{84.1}& \textbf{86.3}& \multicolumn{1}{c|}{\cellcolor[HTML]{CBCEFB}\underline{82.7}} & \underline{78.5}& \textbf{84.1}& \multicolumn{1}{c|}{\cellcolor[HTML]{CBCEFB}\underline{74.3}} & \underline{79.3}& \multicolumn{1}{c|}{\cellcolor[HTML]{CBCEFB}\textbf{79.8}} & \textbf{48.7}& \textbf{58.5}& \textbf{53.0}                      \\ \hline
\end{tabular}
        }
        \vspace{1mm}
        \label{tab:seg}
    \end{minipage}
    \hfill
    \begin{minipage}{0.49\textwidth}
        \centering
        \caption{\textbf{Quantitative results on image editing benchmarks. } The performance with top-1 and top-2 value are denoted in bold and underline.
        }
        \LARGE
        \tablestyle{3.2pt}{1}
        \resizebox{1.0\textwidth}{!}{
        \begin{tabular}{lcc|cccc}
            \toprule
             & & & \multicolumn{4}{c}{\it Emu Edit~\cite{sheynin2023emueditpreciseimage}}  \\
            \multirow{-2}[0]{*}{Method} & \multirow{-2}[0]{*}{Type} & \multirow{-2}[0]{*}{Tasks} & DINO & CLIP-I & CLIP-T & CLIP-DIR \\
            \midrule
            InstructPix2Pix~\cite{flamingo} & Diffusion & Edit only & 0.762 & 0.834 & 0.219 & 0.078 \\
            MagicBrush~\cite{zhang2024magicbrushmanuallyannotateddataset} & Diffusion & Edit only & 0.776 & 0.838 & 0.222 & 0.09\\
            OmniGen~\cite{emu} & Diffusion & Edit only & 0.804 & 0.836 & 0.233 & -\\
            Emu Edit~\cite{sheynin2023emueditpreciseimage} & Diffusion & Edit only & 0.819 & 0.859 & 0.231 & \textbf{0.109}\\
            \midrule
            PUMA~\cite{fang2024pumaempoweringunifiedmllm} & AR & Edit only & 0.785 & 0.846 & 0.270 & -\\
            ILLUME & AR & Und, Gen, Edit & 0.791 & \textbf{0.879} & 0.260 & - \\
            ILLUME+ & AR & Und, Gen, Edit & \underline{0.826} & 0.872 & \underline{0.275} & 0.101 \\
            \rowcolor[HTML]{CBCEFB} 
            FOCUS & AR & Und, Gen, Edit & \textbf{0.831} & \underline{0.876} & \textbf{0.278} & \underline{0.105} \\
            \bottomrule
        \end{tabular}}
        \vspace{1mm}

        \label{tab:editing}
    \end{minipage}
\end{table}
\vspace{-10mm}

\subsection{Referring Segmentation Accuracy}
\label{sec:seg_eval}

We evaluate the referential segmentation performance of \textbf{FOCUS} on four standard benchmarks: RefCOCO, RefCOCO+, RefCOCOg, and gRefCOCO, using mean Intersection-over-Union (mIoU) as the evaluation metric. As shown in table~\ref{tab:seg}, \textbf{FOCUS} achieves competitive or superior performance compared to both segmentation-specific and LVLM-based methods. These results demonstrate the effectiveness of \textbf{FOCUS} in pixel-level target localization and its strong capacity to align complex referring expressions with visual semantics in an end-to-end framework. More empirical evidence is presented in the appendix.

\section{Conclusion}
\label{sec:conclusion}

FOCUS demonstrates robust generalization and strong task performance across multimodal dialogue, referential segmentation, and controlled visual editing. The consistent improvements over baseline models affirm the effectiveness of our unified architecture, which tightly integrates segmentation-aware perception with instruction-guided generation. These results further support the practical value of FOCUS in real-world multimodal interaction scenarios.

{\small
\bibliographystyle{plainnat}
\bibliography{main}

\begin{thebibliography}{87}
\providecommand{\natexlab}[1]{#1}
\providecommand{\url}[1]{\texttt{#1}}
\expandafter\ifx\csname urlstyle\endcsname\relax
  \providecommand{\doi}[1]{doi: #1}\else
  \providecommand{\doi}{doi: \begingroup \urlstyle{rm}\Url}\fi

\bibitem[Alayrac et~al.(2022)Alayrac, Donahue, Luc, Miech, Barr, Hasson, Lenc, Mensch, Millican, Reynolds, et~al.]{flamingo}
Jean-Baptiste Alayrac, Jeff Donahue, Pauline Luc, Antoine Miech, Iain Barr, Yana Hasson, Karel Lenc, Arthur Mensch, Katherine Millican, Malcolm Reynolds, et~al.
\newblock Flamingo: a visual language model for few-shot learning.
\newblock \emph{Advances in Neural Information Processing Systems}, 35:\penalty0 23716--23736, 2022.

\bibitem[Anthropic(2024{\natexlab{a}})]{claude}
Anthropic.
\newblock Claude 3: Next-generation ai models.
\newblock \url{https://www.anthropic.com/news/claude-3}, 2024{\natexlab{a}}.
\newblock Accessed: 2025-05-12.

\bibitem[Anthropic(2024{\natexlab{b}})]{claude3}
Anthropic.
\newblock Claude 3: Next-generation ai models.
\newblock \url{https://www.anthropic.com/news/claude-3}, 2024{\natexlab{b}}.
\newblock Accessed: 2025-05-12.

\bibitem[Bai et~al.(2023)Bai, Bai, Yang, Wang, Tan, Wang, Lin, Zhou, and Zhou]{Qwen-vl}
Jinze Bai, Shuai Bai, Shusheng Yang, Shijie Wang, Sinan Tan, Peng Wang, Junyang Lin, Chang Zhou, and Jingren Zhou.
\newblock Qwen-vl: A versatile vision-language model for understanding, localization, text reading, and beyond.
\newblock \emph{arXiv preprint arXiv:2308.12966}, 2023.

\bibitem[Brooks et~al.(2023)Brooks, Holynski, and Efros]{brooks2023instructpix2pixlearningfollowimage}
Tim Brooks, Aleksander Holynski, and Alexei~A. Efros.
\newblock Instructpix2pix: Learning to follow image editing instructions, 2023.
\newblock URL \url{https://arxiv.org/abs/2211.09800}.

\bibitem[Byeon et~al.(2022)Byeon, Park, Kim, Lee, Baek, and Kim]{kakaobrain2022coyo-700m}
Minwoo Byeon, Beomhee Park, Haecheon Kim, Sungjun Lee, Woonhyuk Baek, and Saehoon Kim.
\newblock Coyo-700m: Image-text pair dataset.
\newblock \url{https://github.com/kakaobrain/coyo-dataset}, 2022.

\bibitem[Caelles et~al.(2018)Caelles, Montes, Maninis, Chen, Gool, Perazzi, and Pont-Tuset]{caelles20182018davischallengevideo}
Sergi Caelles, Alberto Montes, Kevis-Kokitsi Maninis, Yuhua Chen, Luc~Van Gool, Federico Perazzi, and Jordi Pont-Tuset.
\newblock The 2018 davis challenge on video object segmentation, 2018.
\newblock URL \url{https://arxiv.org/abs/1803.00557}.

\bibitem[Chen et~al.(2023{\natexlab{a}})Chen, Yu, Ge, Yao, Xie, Wu, Wang, Kwok, Luo, Lu, and Li]{chen2023pixartalphafasttrainingdiffusion}
Junsong Chen, Jincheng Yu, Chongjian Ge, Lewei Yao, Enze Xie, Yue Wu, Zhongdao Wang, James Kwok, Ping Luo, Huchuan Lu, and Zhenguo Li.
\newblock Pixart-$\alpha$: Fast training of diffusion transformer for photorealistic text-to-image synthesis, 2023{\natexlab{a}}.
\newblock URL \url{https://arxiv.org/abs/2310.00426}.

\bibitem[Chen et~al.(2025{\natexlab{a}})Chen, Gou, Huang, Liu, Tan, Xu, Wang, Zhu, Zeng, Yang, Wang, Xiang, Li, Bai, Han, Li, Jin, Xie, Zhang, Kwok, Zhao, Liang, Yeung, Chen, Li, Zhang, Liu, Yao, Hong, Hou, and Xu]{chen2025emovaempoweringlanguagemodels}
Kai Chen, Yunhao Gou, Runhui Huang, Zhili Liu, Daxin Tan, Jing Xu, Chunwei Wang, Yi~Zhu, Yihan Zeng, Kuo Yang, Dingdong Wang, Kun Xiang, Haoyuan Li, Haoli Bai, Jianhua Han, Xiaohui Li, Weike Jin, Nian Xie, Yu~Zhang, James~T. Kwok, Hengshuang Zhao, Xiaodan Liang, Dit-Yan Yeung, Xiao Chen, Zhenguo Li, Wei Zhang, Qun Liu, Jun Yao, Lanqing Hong, Lu~Hou, and Hang Xu.
\newblock Emova: Empowering language models to see, hear and speak with vivid emotions, 2025{\natexlab{a}}.
\newblock URL \url{https://arxiv.org/abs/2409.18042}.

\bibitem[Chen et~al.(2023{\natexlab{b}})Chen, Li, Dong, Zhang, He, Wang, Zhao, and Lin]{chen2023sharegpt4vimprovinglargemultimodal}
Lin Chen, Jinsong Li, Xiaoyi Dong, Pan Zhang, Conghui He, Jiaqi Wang, Feng Zhao, and Dahua Lin.
\newblock Sharegpt4v: Improving large multi-modal models with better captions, 2023{\natexlab{b}}.
\newblock URL \url{https://arxiv.org/abs/2311.12793}.

\bibitem[Chen et~al.(2023{\natexlab{c}})Chen, Spiridonova, Yang, Gao, and Li]{chen2023llavainteractiveallinonedemoimage}
Wei-Ge Chen, Irina Spiridonova, Jianwei Yang, Jianfeng Gao, and Chunyuan Li.
\newblock Llava-interactive: An all-in-one demo for image chat, segmentation, generation and editing, 2023{\natexlab{c}}.
\newblock URL \url{https://arxiv.org/abs/2311.00571}.

\bibitem[Chen et~al.(2023{\natexlab{d}})Chen, Spiridonova, Yang, Gao, and Li]{llavainteractive}
Wei-Ge Chen, Irina Spiridonova, Jianwei Yang, Jianfeng Gao, and Chunyuan Li.
\newblock Llava-interactive: An all-in-one demo for image chat, segmentation, generation and editing.
\newblock \emph{arXiv preprint arXiv:2311.00571}, 2023{\natexlab{d}}.

\bibitem[Chen et~al.(2014)Chen, Mottaghi, Liu, Fidler, Urtasun, and Yuille]{chen2014detect}
Xianjie Chen, Roozbeh Mottaghi, Xiaobai Liu, Sanja Fidler, Raquel Urtasun, and Alan Yuille.
\newblock Detect what you can: Detecting and representing objects using holistic models and body parts.
\newblock In \emph{Proceedings of the IEEE conference on computer vision and pattern recognition}, pages 1971--1978, 2014.

\bibitem[Chen et~al.(2025{\natexlab{b}})Chen, Wu, Liu, Pan, Liu, Xie, Yu, and Ruan]{chen2025janusprounifiedmultimodalunderstanding}
Xiaokang Chen, Zhiyu Wu, Xingchao Liu, Zizheng Pan, Wen Liu, Zhenda Xie, Xingkai Yu, and Chong Ruan.
\newblock Janus-pro: Unified multimodal understanding and generation with data and model scaling, 2025{\natexlab{b}}.
\newblock URL \url{https://arxiv.org/abs/2501.17811}.

\bibitem[Cheng et~al.(2022)Cheng, Misra, Schwing, Kirillov, and Girdhar]{cheng2022maskedattentionmasktransformeruniversal}
Bowen Cheng, Ishan Misra, Alexander~G. Schwing, Alexander Kirillov, and Rohit Girdhar.
\newblock Masked-attention mask transformer for universal image segmentation, 2022.
\newblock URL \url{https://arxiv.org/abs/2112.01527}.

\bibitem[Couairon et~al.(2022)Couairon, Verbeek, Schwenk, and Cord]{couairon2022diffeditdiffusionbasedsemanticimage}
Guillaume Couairon, Jakob Verbeek, Holger Schwenk, and Matthieu Cord.
\newblock Diffedit: Diffusion-based semantic image editing with mask guidance, 2022.
\newblock URL \url{https://arxiv.org/abs/2210.11427}.

\bibitem[Dai et~al.(2024)Dai, Li, Li, Tiong, Zhao, Wang, Li, Fung, and Hoi]{Instructblip}
Wenliang Dai, Junnan Li, Dongxu Li, Anthony Meng~Huat Tiong, Junqi Zhao, Weisheng Wang, Boyang Li, Pascale~N Fung, and Steven Hoi.
\newblock Instructblip: Towards general-purpose vision-language models with instruction tuning.
\newblock \emph{Advances in Neural Information Processing Systems}, 36, 2024.

\bibitem[Deitke et~al.(2024)Deitke, Clark, Lee, Tripathi, Yang, Park, Salehi, Muennighoff, Lo, Soldaini, Lu, Anderson, Bransom, Ehsani, Ngo, Chen, Patel, Yatskar, Callison-Burch, Head, Hendrix, Bastani, VanderBilt, Lambert, Chou, Chheda, Sparks, Skjonsberg, Schmitz, Sarnat, Bischoff, Walsh, Newell, Wolters, Gupta, Zeng, Borchardt, Groeneveld, Nam, Lebrecht, Wittlif, Schoenick, Michel, Krishna, Weihs, Smith, Hajishirzi, Girshick, Farhadi, and Kembhavi]{deitke2024molmopixmoopenweights}
Matt Deitke, Christopher Clark, Sangho Lee, Rohun Tripathi, Yue Yang, Jae~Sung Park, Mohammadreza Salehi, Niklas Muennighoff, Kyle Lo, Luca Soldaini, Jiasen Lu, Taira Anderson, Erin Bransom, Kiana Ehsani, Huong Ngo, YenSung Chen, Ajay Patel, Mark Yatskar, Chris Callison-Burch, Andrew Head, Rose Hendrix, Favyen Bastani, Eli VanderBilt, Nathan Lambert, Yvonne Chou, Arnavi Chheda, Jenna Sparks, Sam Skjonsberg, Michael Schmitz, Aaron Sarnat, Byron Bischoff, Pete Walsh, Chris Newell, Piper Wolters, Tanmay Gupta, Kuo-Hao Zeng, Jon Borchardt, Dirk Groeneveld, Crystal Nam, Sophie Lebrecht, Caitlin Wittlif, Carissa Schoenick, Oscar Michel, Ranjay Krishna, Luca Weihs, Noah~A. Smith, Hannaneh Hajishirzi, Ross Girshick, Ali Farhadi, and Aniruddha Kembhavi.
\newblock Molmo and pixmo: Open weights and open data for state-of-the-art vision-language models, 2024.
\newblock URL \url{https://arxiv.org/abs/2409.17146}.

\bibitem[Fang et~al.(2024)Fang, Duan, Wang, Li, Tian, Zeng, Zhao, Dai, Li, and Liu]{fang2024pumaempoweringunifiedmllm}
Rongyao Fang, Chengqi Duan, Kun Wang, Hao Li, Hao Tian, Xingyu Zeng, Rui Zhao, Jifeng Dai, Hongsheng Li, and Xihui Liu.
\newblock Puma: Empowering unified mllm with multi-granular visual generation, 2024.
\newblock URL \url{https://arxiv.org/abs/2410.13861}.

\bibitem[Gou et~al.(2024)Gou, Shao, Gong, Shen, Yang, Huang, Duan, and Chen]{gou2024toratoolintegratedreasoningagent}
Zhibin Gou, Zhihong Shao, Yeyun Gong, Yelong Shen, Yujiu Yang, Minlie Huang, Nan Duan, and Weizhu Chen.
\newblock Tora: A tool-integrated reasoning agent for mathematical problem solving, 2024.
\newblock URL \url{https://arxiv.org/abs/2309.17452}.

\bibitem[He et~al.(2022)He, Yang, Yang, Kortylewski, Yuan, Chen, Liu, Yang, Yu, and Yuille]{he2022partimagenet}
Ju~He, Shuo Yang, Shaokang Yang, Adam Kortylewski, Xiaoding Yuan, Jie-Neng Chen, Shuai Liu, Cheng Yang, Qihang Yu, and Alan Yuille.
\newblock Partimagenet: A large, high-quality dataset of parts.
\newblock In \emph{European Conference on Computer Vision}, pages 128--145. Springer, 2022.

\bibitem[Huang et~al.(2025{\natexlab{a}})Huang, Wang, Yang, Lu, Yuan, Han, Hou, Zhang, Hong, Zhao, and Xu]{huang2025illumeilluminatingunifiedmllm}
Runhui Huang, Chunwei Wang, Junwei Yang, Guansong Lu, Yunlong Yuan, Jianhua Han, Lu~Hou, Wei Zhang, Lanqing Hong, Hengshuang Zhao, and Hang Xu.
\newblock Illume+: Illuminating unified mllm with dual visual tokenization and diffusion refinement, 2025{\natexlab{a}}.
\newblock URL \url{https://arxiv.org/abs/2504.01934}.

\bibitem[Huang et~al.(2025{\natexlab{b}})Huang, Cheng, Liu, Hao, Song, Xu, Yang, Liu, Zhang, Chai, Yuan, Zhang, Fu, Liu, Zhang, Wang, Qi, Xu, and Chu]{huang2025opencoderopencookbooktoptier}
Siming Huang, Tianhao Cheng, J.~K. Liu, Jiaran Hao, Liuyihan Song, Yang Xu, J.~Yang, Jiaheng Liu, Chenchen Zhang, Linzheng Chai, Ruifeng Yuan, Zhaoxiang Zhang, Jie Fu, Qian Liu, Ge~Zhang, Zili Wang, Yuan Qi, Yinghui Xu, and Wei Chu.
\newblock Opencoder: The open cookbook for top-tier code large language models, 2025{\natexlab{b}}.
\newblock URL \url{https://arxiv.org/abs/2411.04905}.

\bibitem[Kazemzadeh et~al.(2014)Kazemzadeh, Ordonez, Matten, and Berg]{refclef}
Sahar Kazemzadeh, Vicente Ordonez, Mark Matten, and Tamara Berg.
\newblock Referitgame: Referring to objects in photographs of natural scenes.
\newblock In \emph{Proceedings of the 2014 Conference on Empirical Methods in Natural Language Processing (EMNLP)}, Jan 2014.
\newblock \doi{10.3115/v1/d14-1086}.
\newblock URL \url{http://dx.doi.org/10.3115/v1/d14-1086}.

\bibitem[Lai et~al.(2023)Lai, Tian, Chen, Li, Yuan, Liu, and Jia]{lisa}
Xin Lai, Zhuotao Tian, Yukang Chen, Yanwei Li, Yuhui Yuan, Shu Liu, and Jiaya Jia.
\newblock Lisa: Reasoning segmentation via large language model.
\newblock \emph{arXiv preprint arXiv:2308.00692}, 2023.

\bibitem[Li et~al.(2024{\natexlab{a}})Li, Zhang, Guo, Zhang, Li, Zhang, Zhang, Zhang, Li, Liu, and Li]{li2024llavaonevisioneasyvisualtask}
Bo~Li, Yuanhan Zhang, Dong Guo, Renrui Zhang, Feng Li, Hao Zhang, Kaichen Zhang, Peiyuan Zhang, Yanwei Li, Ziwei Liu, and Chunyuan Li.
\newblock Llava-onevision: Easy visual task transfer, 2024{\natexlab{a}}.
\newblock URL \url{https://arxiv.org/abs/2408.03326}.

\bibitem[Li et~al.(2023{\natexlab{a}})Li, Wang, Wang, Ge, Ge, and Shan]{li2023seedbenchbenchmarkingmultimodalllms}
Bohao Li, Rui Wang, Guangzhi Wang, Yuying Ge, Yixiao Ge, and Ying Shan.
\newblock Seed-bench: Benchmarking multimodal llms with generative comprehension, 2023{\natexlab{a}}.
\newblock URL \url{https://arxiv.org/abs/2307.16125}.

\bibitem[Li et~al.()Li, Li, Savarese, and Hoi]{BLIP-2}
Junnan Li, Dongxu Li, Silvio Savarese, and Steven Hoi.
\newblock Blip-2: Bootstrapping language-image pre-training with frozen image encoders and large language models.

\bibitem[Li et~al.(2023{\natexlab{b}})Li, Wang, Xu, Li, Raj, and Lu]{li2023robustreferringvideoobject}
Xiang Li, Jinglu Wang, Xiaohao Xu, Xiao Li, Bhiksha Raj, and Yan Lu.
\newblock Towards robust referring video object segmentation with cyclic relational consensus, 2023{\natexlab{b}}.
\newblock URL \url{https://arxiv.org/abs/2207.01203}.

\bibitem[Li et~al.(2023{\natexlab{c}})Li, Du, Zhou, Wang, Zhao, and Wen]{li2023evaluatingobjecthallucinationlarge}
Yifan Li, Yifan Du, Kun Zhou, Jinpeng Wang, Wayne~Xin Zhao, and Ji-Rong Wen.
\newblock Evaluating object hallucination in large vision-language models, 2023{\natexlab{c}}.
\newblock URL \url{https://arxiv.org/abs/2305.10355}.

\bibitem[Li et~al.(2024{\natexlab{b}})Li, Wang, Cai, Qi, Wang, Zhang, Song, Jiang, Huang, and Wang]{li2024unifiedmllmenablingunifiedrepresentation}
Zhaowei Li, Wei Wang, YiQing Cai, Xu~Qi, Pengyu Wang, Dong Zhang, Hang Song, Botian Jiang, Zhida Huang, and Tao Wang.
\newblock Unifiedmllm: Enabling unified representation for multi-modal multi-tasks with large language model, 2024{\natexlab{b}}.
\newblock URL \url{https://arxiv.org/abs/2408.02503}.

\bibitem[Lian et~al.(2025)Lian, Ding, Ge, Liu, Mao, Li, Pavone, Liu, Darrell, Yala, and Cui]{lian2025anythingdetailedlocalizedimage}
Long Lian, Yifan Ding, Yunhao Ge, Sifei Liu, Hanzi Mao, Boyi Li, Marco Pavone, Ming-Yu Liu, Trevor Darrell, Adam Yala, and Yin Cui.
\newblock Describe anything: Detailed localized image and video captioning, 2025.
\newblock URL \url{https://arxiv.org/abs/2504.16072}.

\bibitem[Lian et~al.(2023)Lian, Goodson, Pentland, Cook, Vong, and "Teknium"]{OpenOrca}
Wing Lian, Bleys Goodson, Eugene Pentland, Austin Cook, Chanvichet Vong, and "Teknium".
\newblock Openorca: An open dataset of gpt augmented flan reasoning traces.
\newblock \url{https://https://huggingface.co/datasets/Open-Orca/OpenOrca}, 2023.

\bibitem[Lin et~al.(2023)Lin, Ye, Zhu, Cui, Ning, Jin, and Yuan]{video-llava}
Bin Lin, Yang Ye, Bin Zhu, Jiaxi Cui, Munan Ning, Peng Jin, and Li~Yuan.
\newblock Video-llava: Learning united visual representation by alignment before projection.
\newblock \emph{arXiv preprint arXiv:2311.10122}, 2023.

\bibitem[Liu et~al.(2023{\natexlab{a}})Liu, Ding, and Jiang]{grefcoco}
Chang Liu, Henghui Ding, and Xudong Jiang.
\newblock Gres: Generalized referring expression segmentation.
\newblock In \emph{Proceedings of the IEEE/CVF Conference on Computer Vision and Pattern Recognition}, pages 23592--23601, 2023{\natexlab{a}}.

\bibitem[Liu et~al.(2025)Liu, Yan, Zaharia, and Abbeel]{liu2025worldmodelmillionlengthvideo}
Hao Liu, Wilson Yan, Matei Zaharia, and Pieter Abbeel.
\newblock World model on million-length video and language with blockwise ringattention, 2025.
\newblock URL \url{https://arxiv.org/abs/2402.08268}.

\bibitem[Liu et~al.(2023{\natexlab{b}})Liu, Li, Li, and Lee]{liu2023improvedllava}
Haotian Liu, Chunyuan Li, Yuheng Li, and Yong~Jae Lee.
\newblock Improved baselines with visual instruction tuning, 2023{\natexlab{b}}.

\bibitem[Liu et~al.(2023{\natexlab{c}})Liu, Li, Wu, and Lee]{liu2023llava}
Haotian Liu, Chunyuan Li, Qingyang Wu, and Yong~Jae Lee.
\newblock Visual instruction tuning, 2023{\natexlab{c}}.

\bibitem[Liu et~al.(2024{\natexlab{a}})Liu, Li, Wu, and Lee]{LLaVA}
Haotian Liu, Chunyuan Li, Qingyang Wu, and Yong~Jae Lee.
\newblock Visual instruction tuning.
\newblock \emph{Advances in neural information processing systems}, 36, 2024{\natexlab{a}}.

\bibitem[Liu et~al.(2023{\natexlab{d}})Liu, Ding, Cai, Zhang, Satzoda, Mahadevan, and Manmatha]{PolyFormer}
Jiang Liu, Hui Ding, Zhaowei Cai, Yuting Zhang, Ravi~Kumar Satzoda, Vijay Mahadevan, and R~Manmatha.
\newblock Polyformer: Referring image segmentation as sequential polygon generation.
\newblock In \emph{Proceedings of the IEEE/CVF Conference on Computer Vision and Pattern Recognition}, pages 18653--18663, 2023{\natexlab{d}}.

\bibitem[Liu et~al.(2024{\natexlab{b}})Liu, Duan, Zhang, Li, Zhang, Zhao, Yuan, Wang, He, Liu, Chen, and Lin]{liu2024mmbenchmultimodalmodelallaround}
Yuan Liu, Haodong Duan, Yuanhan Zhang, Bo~Li, Songyang Zhang, Wangbo Zhao, Yike Yuan, Jiaqi Wang, Conghui He, Ziwei Liu, Kai Chen, and Dahua Lin.
\newblock Mmbench: Is your multi-modal model an all-around player?, 2024{\natexlab{b}}.
\newblock URL \url{https://arxiv.org/abs/2307.06281}.

\bibitem[Liu et~al.(2022)Liu, Mao, Wu, Feichtenhofer, Darrell, and Xie]{liu2022convnet}
Zhuang Liu, Hanzi Mao, Chao-Yuan Wu, Christoph Feichtenhofer, Trevor Darrell, and Saining Xie.
\newblock A convnet for the 2020s.
\newblock In \emph{Proceedings of the IEEE/CVF conference on computer vision and pattern recognition}, pages 11976--11986, 2022.

\bibitem[Lu et~al.(2025)Lu, Tan, Xu, Yao, Qu, Chu, Xu, and Qi]{lu2025scp116khighqualityproblemsolutiondataset}
Dakuan Lu, Xiaoyu Tan, Rui Xu, Tianchu Yao, Chao Qu, Wei Chu, Yinghui Xu, and Yuan Qi.
\newblock Scp-116k: A high-quality problem-solution dataset and a generalized pipeline for automated extraction in the higher education science domain, 2025.
\newblock URL \url{https://arxiv.org/abs/2501.15587}.

\bibitem[Lu et~al.(2022)Lu, Clark, Zellers, Mottaghi, and Kembhavi]{lu2022unifiediounifiedmodelvision}
Jiasen Lu, Christopher Clark, Rowan Zellers, Roozbeh Mottaghi, and Aniruddha Kembhavi.
\newblock Unified-io: A unified model for vision, language, and multi-modal tasks, 2022.
\newblock URL \url{https://arxiv.org/abs/2206.08916}.

\bibitem[Mao et~al.(2016)Mao, Huang, Toshev, Camburu, Yuille, and Murphy]{refcocog}
Junhua Mao, Jonathan Huang, Alexander Toshev, Oana Camburu, Alan Yuille, and Kevin Murphy.
\newblock Generation and comprehension of unambiguous object descriptions.
\newblock In \emph{2016 IEEE Conference on Computer Vision and Pattern Recognition (CVPR)}, Jun 2016.
\newblock \doi{10.1109/cvpr.2016.9}.
\newblock URL \url{http://dx.doi.org/10.1109/cvpr.2016.9}.

\bibitem[OpenAI(2023)]{chatgpt}
OpenAI.
\newblock Chatgpt: Language model (gpt-4).
\newblock \url{https://chat.openai.com}, 2023.
\newblock Accessed: 2025-05-12.

\bibitem[Peebles and Xie(2023)]{peebles2023scalablediffusionmodelstransformers}
William Peebles and Saining Xie.
\newblock Scalable diffusion models with transformers, 2023.
\newblock URL \url{https://arxiv.org/abs/2212.09748}.

\bibitem[Podell et~al.(2023)Podell, English, Lacey, Blattmann, Dockhorn, Müller, Penna, and Rombach]{podell2023sdxlimprovinglatentdiffusion}
Dustin Podell, Zion English, Kyle Lacey, Andreas Blattmann, Tim Dockhorn, Jonas Müller, Joe Penna, and Robin Rombach.
\newblock Sdxl: Improving latent diffusion models for high-resolution image synthesis, 2023.
\newblock URL \url{https://arxiv.org/abs/2307.01952}.

\bibitem[Pont-Tuset et~al.(2020)Pont-Tuset, Uijlings, Changpinyo, Soricut, and Ferrari]{ponttuset2020connectingvisionlanguagelocalized}
Jordi Pont-Tuset, Jasper Uijlings, Soravit Changpinyo, Radu Soricut, and Vittorio Ferrari.
\newblock Connecting vision and language with localized narratives, 2020.
\newblock URL \url{https://arxiv.org/abs/1912.03098}.

\bibitem[Ramanathan et~al.(2023)Ramanathan, Kalia, Petrovic, Wen, Zheng, Guo, Wang, Marquez, Kovvuri, Kadian, et~al.]{ramanathan2023paco}
Vignesh Ramanathan, Anmol Kalia, Vladan Petrovic, Yi~Wen, Baixue Zheng, Baishan Guo, Rui Wang, Aaron Marquez, Rama Kovvuri, Abhishek Kadian, et~al.
\newblock Paco: Parts and attributes of common objects.
\newblock In \emph{Proceedings of the IEEE/CVF Conference on Computer Vision and Pattern Recognition}, pages 7141--7151, 2023.

\bibitem[Ren et~al.(2023)Ren, Huang, Wei, Zhao, Fu, Feng, and Jin]{pixellm}
Zhongwei Ren, Zhicheng Huang, Yunchao Wei, Yao Zhao, Dongmei Fu, Jiashi Feng, and Xiaojie Jin.
\newblock Pixellm: Pixel reasoning with large multimodal model.
\newblock \emph{arXiv preprint arXiv:2312.02228}, 2023.

\bibitem[Rombach et~al.(2022)Rombach, Blattmann, Lorenz, Esser, and Ommer]{Rombach_2022_CVPR}
Robin Rombach, Andreas Blattmann, Dominik Lorenz, Patrick Esser, and Bj\"orn Ommer.
\newblock High-resolution image synthesis with latent diffusion models.
\newblock In \emph{Proceedings of the IEEE/CVF Conference on Computer Vision and Pattern Recognition (CVPR)}, pages 10684--10695, June 2022.

\bibitem[Seo et~al.(2020)Seo, Lee, and Han]{10.1007/978-3-030-58555-6_13}
Seonguk Seo, Joon-Young Lee, and Bohyung Han.
\newblock Urvos: Unified referring video object segmentation network with a large-scale benchmark.
\newblock In Andrea Vedaldi, Horst Bischof, Thomas Brox, and Jan-Michael Frahm, editors, \emph{Computer Vision -- ECCV 2020}, pages 208--223, Cham, 2020. Springer International Publishing.
\newblock ISBN 978-3-030-58555-6.

\bibitem[Sheynin et~al.(2023)Sheynin, Polyak, Singer, Kirstain, Zohar, Ashual, Parikh, and Taigman]{sheynin2023emueditpreciseimage}
Shelly Sheynin, Adam Polyak, Uriel Singer, Yuval Kirstain, Amit Zohar, Oron Ashual, Devi Parikh, and Yaniv Taigman.
\newblock Emu edit: Precise image editing via recognition and generation tasks, 2023.
\newblock URL \url{https://arxiv.org/abs/2311.10089}.

\bibitem[Shi et~al.(2024)Shi, Wang, and Huang]{shi2024seededitalignimageregeneration}
Yichun Shi, Peng Wang, and Weilin Huang.
\newblock Seededit: Align image re-generation to image editing, 2024.
\newblock URL \url{https://arxiv.org/abs/2411.06686}.

\bibitem[Sun et~al.(2023)Sun, Yu, Cui, Zhang, Zhang, Wang, Gao, Liu, Huang, and Wang]{emu}
Quan Sun, Qiying Yu, Yufeng Cui, Fan Zhang, Xiaosong Zhang, Yueze Wang, Hongcheng Gao, Jingjing Liu, Tiejun Huang, and Xinlong Wang.
\newblock Emu: Generative pretraining in multimodality.
\newblock \emph{arXiv preprint arXiv:2307.05222}, 2023.

\bibitem[Sun et~al.(2024)Sun, Cui, Zhang, Zhang, Yu, Wang, Rao, Liu, Huang, and Wang]{emu2}
Quan Sun, Yufeng Cui, Xiaosong Zhang, Fan Zhang, Qiying Yu, Yueze Wang, Yongming Rao, Jingjing Liu, Tiejun Huang, and Xinlong Wang.
\newblock Generative multimodal models are in-context learners.
\newblock In \emph{Proceedings of the IEEE/CVF Conference on Computer Vision and Pattern Recognition}, pages 14398--14409, 2024.

\bibitem[Team(2025)]{chameleonteam2025chameleonmixedmodalearlyfusionfoundation}
Chameleon Team.
\newblock Chameleon: Mixed-modal early-fusion foundation models, 2025.
\newblock URL \url{https://arxiv.org/abs/2405.09818}.

\bibitem[Teknium(2023)]{OpenHermes2.5}
Teknium.
\newblock Openhermes 2.5: An open dataset of synthetic data for generalist llm assistants, 2023.
\newblock URL \url{https://huggingface.co/datasets/teknium/OpenHermes-2.5}.

\bibitem[Voigtlaender et~al.(2023)Voigtlaender, Changpinyo, Pont-Tuset, Soricut, and Ferrari]{voigtlaender2023connectingvisionlanguagevideo}
Paul Voigtlaender, Soravit Changpinyo, Jordi Pont-Tuset, Radu Soricut, and Vittorio Ferrari.
\newblock Connecting vision and language with video localized narratives, 2023.
\newblock URL \url{https://arxiv.org/abs/2302.11217}.

\bibitem[Wang et~al.(2023{\natexlab{a}})Wang, Zhang, Birsak, and Wonka]{wang2023instructeditimprovingautomaticmasks}
Qian Wang, Biao Zhang, Michael Birsak, and Peter Wonka.
\newblock Instructedit: Improving automatic masks for diffusion-based image editing with user instructions, 2023{\natexlab{a}}.
\newblock URL \url{https://arxiv.org/abs/2305.18047}.

\bibitem[Wang et~al.(2023{\natexlab{b}})Wang, Lv, Yu, Hong, Qi, Wang, Ji, Yang, Zhao, Song, Xu, Xu, Li, Dong, Ding, and Tang]{CogVLM}
Weihan Wang, Qingsong Lv, Wenmeng Yu, Wenyi Hong, Ji~Qi, Yan Wang, Junhui Ji, Zhuoyi Yang, Lei Zhao, Xixuan Song, Jiazheng Xu, Bin Xu, Juanzi Li, Yuxiao Dong, Ming Ding, and Jie Tang.
\newblock Cogvlm: Visual expert for pretrained language models.
\newblock Nov 2023{\natexlab{b}}.

\bibitem[Wang et~al.(2024)Wang, Zhang, Luo, Sun, Cui, Wang, Zhang, Wang, Li, Yu, Zhao, Ao, Min, Li, Wu, Zhao, Zhang, Wang, Liu, He, Yang, Liu, Lin, Huang, and Wang]{wang2024emu3nexttokenpredictionneed}
Xinlong Wang, Xiaosong Zhang, Zhengxiong Luo, Quan Sun, Yufeng Cui, Jinsheng Wang, Fan Zhang, Yueze Wang, Zhen Li, Qiying Yu, Yingli Zhao, Yulong Ao, Xuebin Min, Tao Li, Boya Wu, Bo~Zhao, Bowen Zhang, Liangdong Wang, Guang Liu, Zheqi He, Xi~Yang, Jingjing Liu, Yonghua Lin, Tiejun Huang, and Zhongyuan Wang.
\newblock Emu3: Next-token prediction is all you need, 2024.
\newblock URL \url{https://arxiv.org/abs/2409.18869}.

\bibitem[Webster et~al.(2023)Webster, Rabin, Simon, and Jurie]{webster2023deduplicationlaion2b}
Ryan Webster, Julien Rabin, Loic Simon, and Frederic Jurie.
\newblock On the de-duplication of laion-2b, 2023.
\newblock URL \url{https://arxiv.org/abs/2303.12733}.

\bibitem[Wei et~al.(2024)Wei, Zhong, Tan, Liu, Zhao, Hu, and Yang]{wei2024hyperseguniversalvisualsegmentation}
Cong Wei, Yujie Zhong, Haoxian Tan, Yong Liu, Zheng Zhao, Jie Hu, and Yujiu Yang.
\newblock Hyperseg: Towards universal visual segmentation with large language model, 2024.
\newblock URL \url{https://arxiv.org/abs/2411.17606}.

\bibitem[Wei et~al.(2025)Wei, Xiong, Ren, Du, Zhang, and Chen]{wei2025omnieditbuildingimageediting}
Cong Wei, Zheyang Xiong, Weiming Ren, Xinrun Du, Ge~Zhang, and Wenhu Chen.
\newblock Omniedit: Building image editing generalist models through specialist supervision, 2025.
\newblock URL \url{https://arxiv.org/abs/2411.07199}.

\bibitem[Wu et~al.(2024)Wu, Chen, Wu, Ma, Liu, Pan, Liu, Xie, Yu, Ruan, and Luo]{wu2024janusdecouplingvisualencoding}
Chengyue Wu, Xiaokang Chen, Zhiyu Wu, Yiyang Ma, Xingchao Liu, Zizheng Pan, Wen Liu, Zhenda Xie, Xingkai Yu, Chong Ruan, and Ping Luo.
\newblock Janus: Decoupling visual encoding for unified multimodal understanding and generation, 2024.
\newblock URL \url{https://arxiv.org/abs/2410.13848}.

\bibitem[Wu et~al.(2025)Wu, Zhang, Chen, Tang, Li, Fang, Zhu, Xie, Yin, Yi, Han, and Lu]{wu2025vilauunifiedfoundationmodel}
Yecheng Wu, Zhuoyang Zhang, Junyu Chen, Haotian Tang, Dacheng Li, Yunhao Fang, Ligeng Zhu, Enze Xie, Hongxu Yin, Li~Yi, Song Han, and Yao Lu.
\newblock Vila-u: a unified foundation model integrating visual understanding and generation, 2025.
\newblock URL \url{https://arxiv.org/abs/2409.04429}.

\bibitem[Xie et~al.(2024)Xie, Mao, Bai, Zhang, Wang, Lin, Gu, Chen, Yang, and Shou]{xie2024showosingletransformerunify}
Jinheng Xie, Weijia Mao, Zechen Bai, David~Junhao Zhang, Weihao Wang, Kevin~Qinghong Lin, Yuchao Gu, Zhijie Chen, Zhenheng Yang, and Mike~Zheng Shou.
\newblock Show-o: One single transformer to unify multimodal understanding and generation, 2024.
\newblock URL \url{https://arxiv.org/abs/2408.12528}.

\bibitem[Xu et~al.(2024)Xu, Jiang, Niu, Deng, Poovendran, Choi, and Lin]{xu2024magpiealignmentdatasynthesis}
Zhangchen Xu, Fengqing Jiang, Luyao Niu, Yuntian Deng, Radha Poovendran, Yejin Choi, and Bill~Yuchen Lin.
\newblock Magpie: Alignment data synthesis from scratch by prompting aligned llms with nothing, 2024.
\newblock URL \url{https://arxiv.org/abs/2406.08464}.

\bibitem[Yan et~al.(2024)Yan, Wang, Yan, Jiang, Hu, Kang, Xie, and Gavves]{yan2024visareasoningvideoobject}
Cilin Yan, Haochen Wang, Shilin Yan, Xiaolong Jiang, Yao Hu, Guoliang Kang, Weidi Xie, and Efstratios Gavves.
\newblock Visa: Reasoning video object segmentation via large language models, 2024.
\newblock URL \url{https://arxiv.org/abs/2407.11325}.

\bibitem[Yang et~al.(2024)Yang, Yang, Zhang, Hui, Zheng, Yu, Li, Liu, Huang, Wei, et~al.]{yang2024qwen2}
An~Yang, Baosong Yang, Beichen Zhang, Binyuan Hui, Bo~Zheng, Bowen Yu, Chengyuan Li, Dayiheng Liu, Fei Huang, Haoran Wei, et~al.
\newblock Qwen2. 5 technical report.
\newblock \emph{arXiv preprint arXiv:2412.15115}, 2024.

\bibitem[Yang et~al.(2019)Yang, Fan, and Xu]{yang2019videoinstancesegmentation}
Linjie Yang, Yuchen Fan, and Ning Xu.
\newblock Video instance segmentation, 2019.
\newblock URL \url{https://arxiv.org/abs/1905.04804}.

\bibitem[Yang et~al.(2023)Yang, Qu, Lai, Tian, Peng, Liu, and Jia]{lisa++}
Senqiao Yang, Tianyuan Qu, Xin Lai, Zhuotao Tian, Bohao Peng, Shu Liu, and Jiaya Jia.
\newblock An improved baseline for reasoning segmentation with large language model.
\newblock \emph{arXiv preprint arXiv:2312.17240}, 2023.

\bibitem[Yang et~al.(2022)Yang, Wang, Tang, Chen, Zhao, and Torr]{lavt}
Zhao Yang, Jiaqi Wang, Yansong Tang, Kai Chen, Hengshuang Zhao, and Philip~HS Torr.
\newblock Lavt: Language-aware vision transformer for referring image segmentation.
\newblock In \emph{Proceedings of the IEEE/CVF Conference on Computer Vision and Pattern Recognition}, pages 18155--18165, 2022.

\bibitem[You et~al.(2023)You, Zhang, Gan, Du, Zhang, Wang, Cao, Chang, and Yang]{Ferret}
Haoxuan You, Haotian Zhang, Zhe Gan, Xianzhi Du, Bowen Zhang, Zirui Wang, Liangliang Cao, Shih-Fu Chang, and Yinfei Yang.
\newblock Ferret: Refer and ground anything anywhere at any granularity.
\newblock Oct 2023.

\bibitem[Yu et~al.(2016)Yu, Poirson, Yang, Berg, and Berg]{refcoco}
Licheng Yu, Patrick Poirson, Shan Yang, Alexander~C Berg, and Tamara~L Berg.
\newblock Modeling context in referring expressions.
\newblock In \emph{Computer Vision--ECCV 2016: 14th European Conference, Amsterdam, The Netherlands, October 11-14, 2016, Proceedings, Part II 14}, pages 69--85. Springer, 2016.

\bibitem[Yu et~al.(2024{\natexlab{a}})Yu, Chow, Yue, Pan, Wu, Wan, Li, Tang, Zhang, and Zhuang]{yu2024anyedit}
Qifan Yu, Wei Chow, Zhongqi Yue, Kaihang Pan, Yang Wu, Xiaoyang Wan, Juncheng Li, Siliang Tang, Hanwang Zhang, and Yueting Zhuang.
\newblock Anyedit: Mastering unified high-quality image editing for any idea.
\newblock \emph{arXiv preprint arXiv:2411.15738}, 2024{\natexlab{a}}.

\bibitem[Yu et~al.(2024{\natexlab{b}})Yu, Yang, Li, Wang, Lin, Liu, Wang, and Wang]{yu2024mmvetevaluatinglargemultimodal}
Weihao Yu, Zhengyuan Yang, Linjie Li, Jianfeng Wang, Kevin Lin, Zicheng Liu, Xinchao Wang, and Lijuan Wang.
\newblock Mm-vet: Evaluating large multimodal models for integrated capabilities, 2024{\natexlab{b}}.
\newblock URL \url{https://arxiv.org/abs/2308.02490}.

\bibitem[Yue et~al.(2024)Yue, Ni, Zhang, Zheng, Liu, Zhang, Stevens, Jiang, Ren, Sun, Wei, Yu, Yuan, Sun, Yin, Zheng, Yang, Liu, Huang, Sun, Su, and Chen]{yue2024mmmumassivemultidisciplinemultimodal}
Xiang Yue, Yuansheng Ni, Kai Zhang, Tianyu Zheng, Ruoqi Liu, Ge~Zhang, Samuel Stevens, Dongfu Jiang, Weiming Ren, Yuxuan Sun, Cong Wei, Botao Yu, Ruibin Yuan, Renliang Sun, Ming Yin, Boyuan Zheng, Zhenzhu Yang, Yibo Liu, Wenhao Huang, Huan Sun, Yu~Su, and Wenhu Chen.
\newblock Mmmu: A massive multi-discipline multimodal understanding and reasoning benchmark for expert agi, 2024.
\newblock URL \url{https://arxiv.org/abs/2311.16502}.

\bibitem[Zellers et~al.()Zellers, Bisk, Farhadi, Choi, and Allen]{VCR}
Rowan Zellers, Yonatan Bisk, Ali Farhadi, Yejin Choi, and Paul Allen.
\newblock From recognition to cognition: Visual commonsense reasoning.

\bibitem[Zhang et~al.(2024{\natexlab{a}})Zhang, Mo, Chen, Sun, and Su]{zhang2024magicbrushmanuallyannotateddataset}
Kai Zhang, Lingbo Mo, Wenhu Chen, Huan Sun, and Yu~Su.
\newblock Magicbrush: A manually annotated dataset for instruction-guided image editing, 2024{\natexlab{a}}.
\newblock URL \url{https://arxiv.org/abs/2306.10012}.

\bibitem[Zhang et~al.(2024{\natexlab{b}})Zhang, Ma, Zhang, and Bai]{zhang2024psalmpixelwisesegmentationlarge}
Zheng Zhang, Yeyao Ma, Enming Zhang, and Xiang Bai.
\newblock Psalm: Pixelwise segmentation with large multi-modal model, 2024{\natexlab{b}}.
\newblock URL \url{https://arxiv.org/abs/2403.14598}.

\bibitem[Zhao et~al.(2024)Zhao, Ma, Chen, Si, Wu, An, Yu, Zhang, Li, and Chang]{zhao2024ultraeditinstructionbasedfinegrainedimage}
Haozhe Zhao, Xiaojian Ma, Liang Chen, Shuzheng Si, Rujie Wu, Kaikai An, Peiyu Yu, Minjia Zhang, Qing Li, and Baobao Chang.
\newblock Ultraedit: Instruction-based fine-grained image editing at scale, 2024.
\newblock URL \url{https://arxiv.org/abs/2407.05282}.

\bibitem[Zheng et~al.(2022)Zheng, Vuong, Cai, and Phung]{zheng2022movqmodulatingquantizedvectors}
Chuanxia Zheng, Long~Tung Vuong, Jianfei Cai, and Dinh Phung.
\newblock Movq: Modulating quantized vectors for high-fidelity image generation, 2022.
\newblock URL \url{https://arxiv.org/abs/2209.09002}.

\bibitem[Zheng et~al.(2024)Zheng, Peng, Yang, Shen, Li, Liu, Zhou, Li, and You]{zheng2024opensorademocratizingefficientvideo}
Zangwei Zheng, Xiangyu Peng, Tianji Yang, Chenhui Shen, Shenggui Li, Hongxin Liu, Yukun Zhou, Tianyi Li, and Yang You.
\newblock Open-sora: Democratizing efficient video production for all, 2024.
\newblock URL \url{https://arxiv.org/abs/2412.20404}.

\bibitem[Zhu et~al.(2023)Zhu, Chen, Shen, Li, and Elhoseiny]{zhu2023minigpt4enhancingvisionlanguageunderstanding}
Deyao Zhu, Jun Chen, Xiaoqian Shen, Xiang Li, and Mohamed Elhoseiny.
\newblock Minigpt-4: Enhancing vision-language understanding with advanced large language models, 2023.
\newblock URL \url{https://arxiv.org/abs/2304.10592}.

\end{thebibliography}
}

\appendix


\section{Dual Quantizer and Diffusion Decoder Training}
To enable high-quality image generation, FOCUS pretrains dual quantizers with a three-branch visual encoder and subsequently trains a diffusion decoder conditioned on the tokens. This section details the training process for both the quantizers and the diffusion decoder.

\begin{figure}[ht]
    \centering
    \includegraphics[width=1\textwidth]{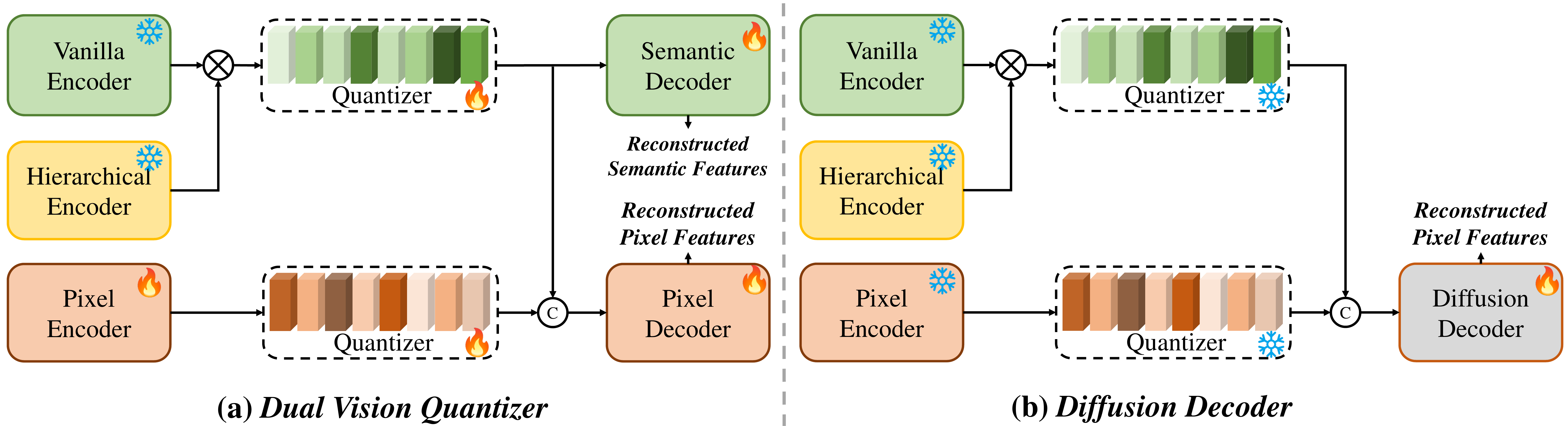}
    \caption{(a) Dual Vision Quantizer: Two separate branches encode semantic and pixel information via quantization and reconstruction. (b) Diffusion Decoder: Reuses pretrained tokens to condition UNet-based denoising for image reconstruction.}
    \label{fig:generation_pretrain}
\end{figure}

\textbf{Dual Quantizer Training.}  As illustrated in Fig.~\ref{fig:generation_pretrain}(a), the semantic quantizer is constructed from two visual branches: a \textit{vanilla encoder} that captures global semantics from low-resolution inputs, and a \textit{hierarchical encoder} that extracts high-resolution spatial features. These two streams cross attention to provide rich semantic representations, which are quantized via a SimVQ~\cite{zhu2023minigpt4enhancingvisionlanguageunderstanding} module and reconstructed using a lightweight transformer as the semantic decoder.

In parallel, a dedicated \textit{pixel encoder} processes high-resolution images to extract fine-grained, low-level textures. Following the MoVQGAN design~\cite{zheng2022movqmodulatingquantizedvectors}, its features are quantized and decoded via a pixel-level decoder. This branch is supervised using L1, perceptual, and adversarial losses.

Each quantizer maintains a separate codebook: 32,768 entries for semantic tokens and 98,304 entries for pixel tokens as ~\cite{huang2025illumeilluminatingunifiedmllm}. The quantizer is trained progressively from $256 \times 256$ to $512 \times 512$ resolution using a bucketed batching strategy, and optimized on a corpus of 45M diverse image-text pairs.


\textbf{Diffusion decoder training.} After quantizer training, we train a UNet-based diffusion decoder, initialized from SDXL~\cite{podell2023sdxlimprovinglatentdiffusion}, to reconstruct high-resolution images conditioned on the learned discrete tokens. As shown in Fig.~\ref{fig:generation_pretrain}(b), semantic and pixel tokens are embedded into continuous representations and concatenated with noisy latents to guide the denoising process.

To accommodate diverse image shapes, 11 aspect-ratio-specific canvas sizes are predefined: \{1:1, 3:4, 4:3, 2:3, 3:2, 1:2, 2:1, 1:3, 3:1, 1:4, 4:1\}~\cite{huang2025illumeilluminatingunifiedmllm}, and samples with more than 20\% content loss after cropping are discarded. The training proceeds in two stages: the first at $512 \times 512$ resolution and the second at $1024 \times 1024$ for super-resolution refinement. During this stage, all encoders and codebooks are frozen, and only the diffusion decoder is updated.

\begin{figure}
    \centering
    \includegraphics[width=1\textwidth]{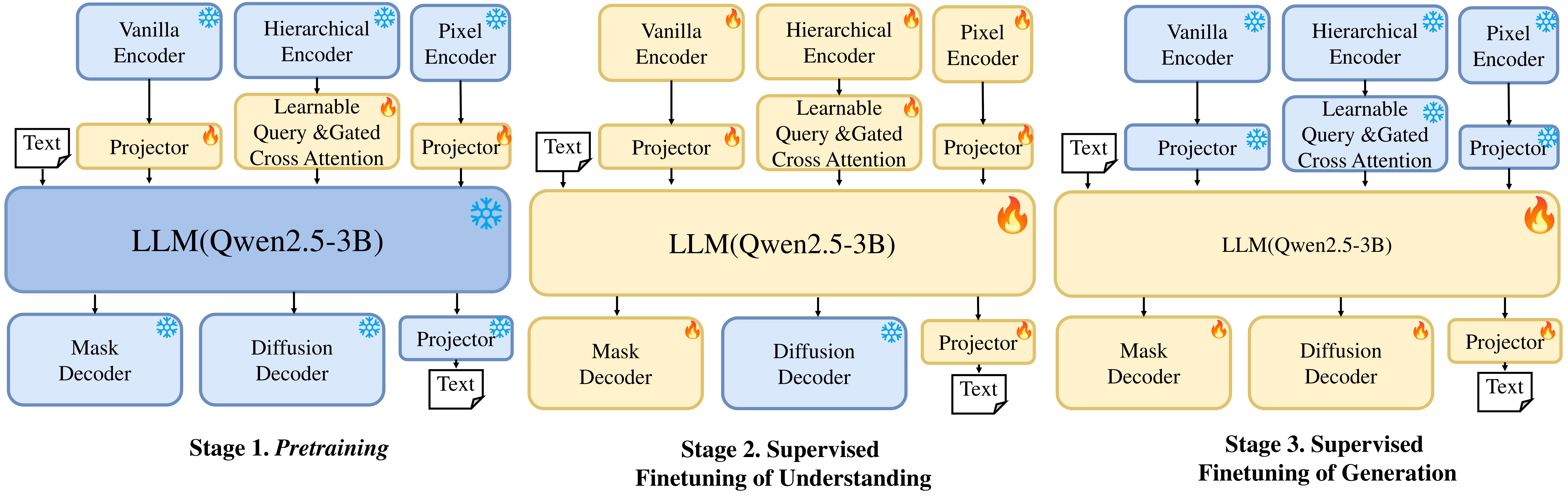}
    \caption{
Three-stage progressive training strategy of FOCUS. Frozen modules are shown in blue, trainable modules in orange. Each stage incrementally activates relevant components to align semantic perception and generative editing.
    }
    \label{fig:diff_stage_v2}
\end{figure}
\section{Progressive Large Vision Language Model Training}


A core challenge in building a unified large vision language model lies in the optimization conflict between high-level semantic understanding in language and low-level visual synthesis in image generation. FOCUS addresses this by adopting a progressive three-stage training paradigm (Fig.~\ref{fig:diff_stage_v2}) that incrementally activates model components to align pixel-level perception and controllable image generation.

\textbf{Visual-Language Projector Warmup.} 
This stage establishes initial alignment between visual features and the large language model (LLM). We train only the projection heads of the vanilla and pixel encoders, along with the learnable queries and gated cross-attention modules in the hierarchical encoder. All backbone encoders, the LLM, the mask decoder is Mask2Former~\cite{cheng2022maskedattentionmasktransformeruniversal}, and the diffusion decoder are frozen. Training is performed on 256$\times$256 image-text pairs using a standard language modeling objective, without any segmentation or generation supervision.


\textbf{Multimodal Pretraining with Segmentation-Aware Alignment.} 
We activate and jointly train the LLM, visual-language adapters, and the mask decoder to enhance multimodal understanding and segmentation capability. The diffusion decoder remains frozen. Supervision includes token-level cross-entropy, segmentation losses (Dice and BCE), and visual reconstruction losses using precomputed features. Training is conducted at both 256$\times$256 and 512$\times$512 resolutions on datasets covering image-text alignment, referring segmentation, and multimodal reasoning.


\textbf{Instruction Tuning for Region-Controlled Editing.} 
This stage focuses on segmentation perception and controllable generation for LVLMs. We instruct tuning the LLM, the mask decoder, and the cross-attention layers in the diffusion decoder, while freezing all visual encoders. Images are processed at resolutions up to 1024$\times$1024. Segmentation outputs are used as spatial guidance and injected into the diffusion decoder’s attention layers. Supervision includes text generation loss, segmentation loss, and L2 loss between denoised outputs and targets.

\begin{table}[ht]
\centering
\caption{Dataset distribution across training stages in FOCUS}
\scalebox{0.70}{
\begin{tabular}{l|l|l|l}
\hline
\textbf{Stage} & \textbf{Number} & \textbf{Task} & \textbf{Source} \\ \hline

\makecell[l]{Stage I: Visual Quantizer and\\ Diffusion Pretraining} 
    & 45M & Image-to-Image & COYO~\cite{kakaobrain2022coyo-700m}, EMOVA~\cite{chen2025emovaempoweringlanguagemodels}, LAION-2B~\cite{webster2023deduplicationlaion2b} \\ \hline

\makecell[l]{Stage II: Visual-Language\\ Projector Warmup} 
    & 30M & Image-to-Text & LLaVA-150K~\cite{liu2023llava, liu2023improvedllava}, COYO, EMOVA-Pretrain \\ \hline

\multirow{8}{*}{\makecell[l]{Stage III: Multimodal Pretraining\\ with Generative and Structural Signals}} 
    & 35M & Image-to-Text \& Editing & UltraEdit~\cite{zhao2024ultraeditinstructionbasedfinegrainedimage}, AnyEdit~\cite{yu2024anyedit}, SEED-Edit~\cite{shi2024seededitalignimageregeneration} \\ \cline{2-4} 
    & \multirow{4}{*}{3M} & \multirow{4}{*}{Segmentation} & RefCOCO-Series~\cite{refcoco, refcocog}, RefClef~\cite{refclef},  \\
    & & & Paco-LVIS~\cite{ramanathan2023paco}, PartImageNet~\cite{he2022partimagenet},  \\
    & & & DAVIS-2017~\cite{caelles20182018davischallengevideo}, Pascal-Part~\cite{chen2014detect}, \\
    & & &  YouTube-VIS2019~\cite{yang2019videoinstancesegmentation} \\ \cline{2-4} 
    & \multirow{3}{*}{5M} & \multirow{3}{*}{Dialog / QA} & Magpie~\cite{xu2024magpiealignmentdatasynthesis}, OpenOrca~\cite{OpenOrca}, \\
    & & & SCP-116K~\cite{lu2025scp116khighqualityproblemsolutiondataset}, OpenHermes~\cite{OpenHermes2.5},  \\ 
    & & & OPC-SFT-Stage1~\cite{huang2025opencoderopencookbooktoptier} \\
    \hline

\multirow{7}{*}{\makecell[l]{Stage IV: Instruction Tuning\\ for Controllable Editing}} 
    & \multirow{3}{*}{7M} & \multirow{3}{*}{Image Editing} & EMOVA-SFT, Pixmo~\cite{deitke2024molmopixmoopenweights}, M4-Instruct~\cite{liu2023improvedllava}, \\
    & & & OmniEdit~\cite{wei2025omnieditbuildingimageediting}, AnyEdit, \\
    & & & UltraEdit, InstructPix2Pix~\cite{brooks2023instructpix2pixlearningfollowimage}, MagPie \\ \cline{2-4} 
    & \multirow{2}{*}{2M} & \multirow{2}{*}{Segmentation} & ReasonSeg~\cite{lisa}, Lisa++ Inst. Seg. \& CoT~\cite{lisa++}, \\
    & & & ReVOS~\cite{yan2024visareasoningvideoobject}, Ref-Youtube-VOS~\cite{10.1007/978-3-030-58555-6_13} \\ \cline{2-4} 
    & 0.5M & Interactive Editing & COCO-Interactive~\cite{zhang2024psalmpixelwisesegmentationlarge} \\ \hline

\end{tabular}}
\label{tab:dataset_stagewise}
\end{table}




\section{Progressive Dataset Structuring}

Each stage in FOCUS adopts dedicated datasets aligned with its training objectives, rather than relying on a uniform corpus across all phases. The dataset distribution and task assignments are summarized in Table~\ref{tab:dataset_stagewise}.

\begin{figure}[ht]
    \centering
    \begin{minipage}{1\textwidth} 
        \centering
        \includegraphics[width=\textwidth]{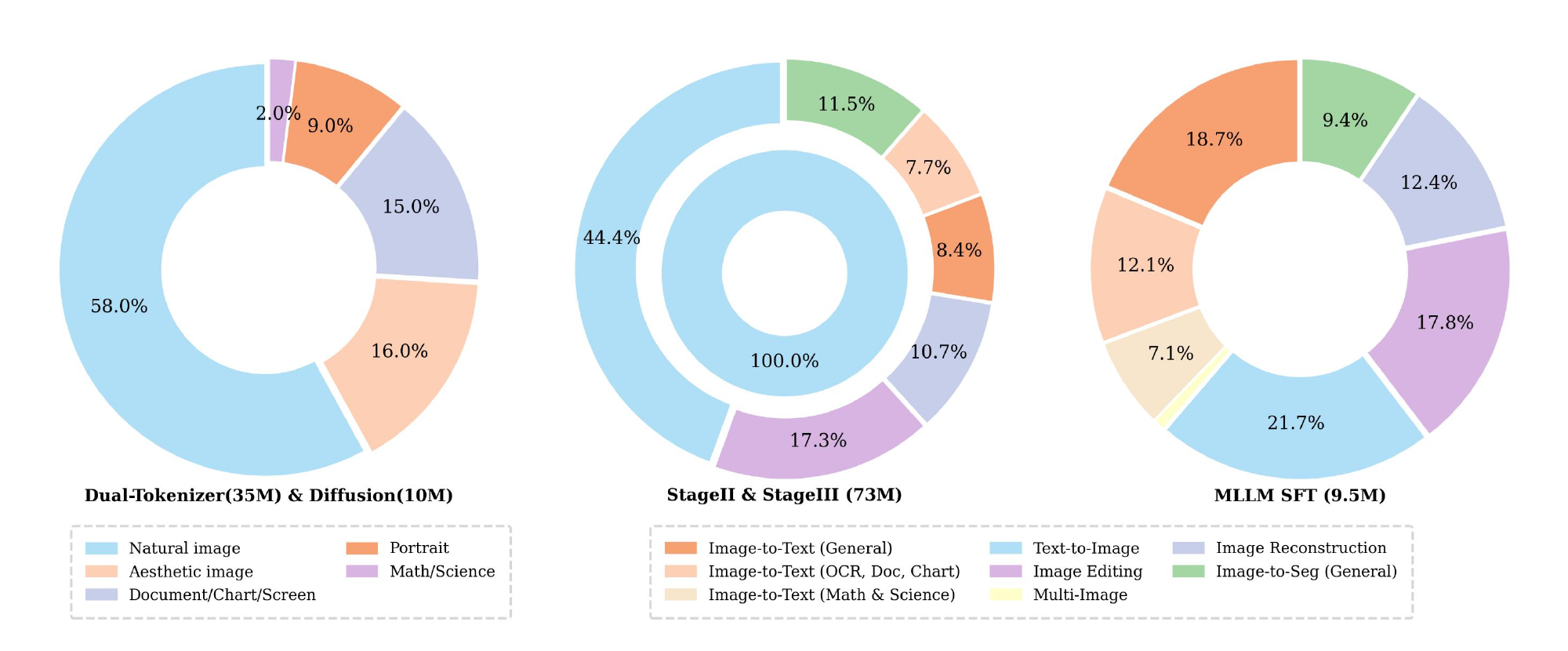}
    \end{minipage}
    \caption{Summary of the data mixture in each stage. Our training data gradually covers a wide range of tasks and various image resoluton.}
    \label{fig:data_composition}
\end{figure}

\textbf{Stage 1: Visual Quantizer and Diffusion Pretraining.} 
This stage focuses on learning discrete representations for both semantic abstraction and fine-grained reconstruction. We utilize 45M image-text pairs from COYO, EMOVA, and LAION-2B. COYO contributes OCR-rich, language-aligned samples; EMOVA provides aesthetic-oriented captions; and LAION-2B enhances domain diversity. These datasets collectively support the training of dual visual quantizers and the diffusion decoder.


\textbf{Stage 2: Visual-Language Adapter Warmup.} To establish foundational alignment between visual embeddings and the language model, we train on 30M image-text pairs from COYO, EMOVA-Pretrain, and LLaVA-150K. These samples span natural scenes, documents, and instruction-following captions. This stage optimizes only the projection heads and adapter modules, using a language modeling loss while keeping all backbone components frozen.


\textbf{Stage 3: Multimodal Pretraining with Generative and Structural Signals.} 
This stage enhances the model’s ability to generate, localize, and reason across modalities. We leverage 35M samples from UltraEdit, SEED-Edit, and AnyEdit for text-guided editing. An additional 3M segmentation annotations are sourced from RefCOCO-series, RefClef, and video datasets like DAVIS-2017 and YouTube-VIS2019. For dialogue and question answering, we include 5M samples from Magpie, OpenOrca, SCP-116K, OpenHermes, and OPC-SFT-Stage1. This stage jointly supervises multimodal reasoning, structural grounding, and fine-grained visual tasks.


\textbf{Stage 4: Instruction Tuning for Controllable Editing.} The final stage fine-tunes the model for spatially guided, instruction-based editing. We use 7M samples from EMOVA-SFT, Pixmo, M4-Instruct, OmniEdit, AnyEdit, UltraEdit, InstructPix2Pix, and Magpie for complex editing tasks. We incorporate 2M segmentation-centric samples from ReasonSeg, Lisa++, and video segmentation sets like RVOS and RefYoutubeVOS. Additionally, COCO-Interactive provides 0.5M samples for interactive region-level editing. This stage improves instruction following, multi-object control, and editing consistency across visual contexts.


\section{Training Configurations Across Different Stages}
\begin{table}[ht]
\centering
\caption{Training hyperparameters across different stages in FOCUS.}
\setlength{\tabcolsep}{4pt}
\renewcommand{\arraystretch}{1.2}
\scalebox{0.6}{
\begin{tabular}{l|c|c|c|c|c}
\toprule
\textbf{Settings} & \textbf{Visual Quantizer} & \textbf{Diffusion Decoder} & \textbf{Projector Warmup} & \textbf{Multimodal Pretraining} & \textbf{Instruction Tuning} \\[-2pt]
& (Tokenizer) & (Image Reconstruction) & (Projector Warmup) & (Seg. Pretrain) &  (Instruction Tuning) \\
\midrule
Learning Rate &
\makecell[c]{1e-4 (semantic) \\ 2e-4 (pixel)} &
2e-5 &
\makecell[c]{1e-3} &
\makecell[c]{2e-5 (Visual encoder, LLM)\\ 1e-3(Mask Decoder)} &
\makecell[c]{2e-5 (Visual encoder, LLM) \\ 2e-6 (Mask Decoder, Diffusion)} \\
\midrule
Batch Size &
256 &
128 &
512 &
128 &
256 \\
Training Steps &
\makecell[c]{136k (pixel) \\ 28k (semantic)} &
220k &
1epoch &
3epoch &
1epoch \\
Image Resolution &
256 to 512 &
512 / 1024 &
256 &
256 / 512 &
512 to 1024 \\
Frozen Modules &
\makecell[c]{Vanilla encoder \\ Hierarchical
Encoder} &
\makecell[c]{All encoders \\ Codebooks} &
\makecell[c]{Visual Encoder, LLM \\  Mask Decoder, Diffusion} &
\makecell[c]{Diffusion} &
\makecell[c]{Visual Encoder} \\
\bottomrule
\end{tabular}}
\label{tab:training-hyperparams}
\end{table}

We adopt stage-specific training configurations to align with the objectives and resolution requirements of each module, as summarized in Table~\ref{tab:training-hyperparams}.

\textbf{Dual Visual Tokenizer} using a fixed learning rate of 1e-4 and batch size of 256. The training follows three progressive resolution stages: 136k steps at 256×256, 28k steps at 512×512. The semantic branch is optimized with a cosine similarity loss, while the pixel branch is trained with a combination of L1, perceptual, and adversarial losses.

\textbf{Diffusion Decoder} using a learning rate of 2e-5 and batch size of 128 for 265k steps. Training employs multi-aspect-ratio cropping up to 1024×1024 resolution. All visual encoders and tokenizers are frozen, and supervision is provided through L2 reconstruction loss.

\textbf{LVLM Training} is conducted in three distinct stages.

\begin{itemize}
  \item \textbf{Projector Warmup}: Only the projection heads and gated visual adapters are trained using learning rates of $1\mathrm{e}{-3}$ (projectors), with a batch size of 512 at $256 \times 256$ resolution for 1 epoch.

  \item \textbf{Multimodal Pretraining}: The visual encoder, large language model and mask decoder are optimized jointly. Phase one uses $256 \times 256$ resolution with a batch size of 256 for 1 epoch, and phase two uses $512 \times 512$ with a batch size of 128 for 2 epochs. A learning rate of $2\mathrm{e}{-5}$ is used throughout, and $1\mathrm{e}{-3}$ is used for the mask decoder.

  \item \textbf{Instruction Tuning}: The LLM, segmentation decoder, and diffusion model’s attention layers are fine-tuned with learning rates of $2\mathrm{e}{-5}$ (LLM) and $2\mathrm{e}{-6}$ (mask decoder, diffusion), with a batch size of 256 for 1 epoch. Training involves input resolutions randomly sampled from $512 \times 512$ to $1024 \times 1024$ using bucketed cropping.
\end{itemize}





\section{Metrics.} 
We evaluate FOCUS across three core tasks using standard benchmarks and metrics. For multimodal understanding, we report accuracy on POPE~\cite{li2023evaluatingobjecthallucinationlarge}, MM-Vet~\cite{yu2024mmvetevaluatinglargemultimodal}, MMBench~\cite{liu2024mmbenchmultimodalmodelallaround}, SEED~\cite{li2023seedbenchbenchmarkingmultimodalllms}, and MMMU~\cite{yue2024mmmumassivemultidisciplinemultimodal} to assess the model’s ability in vision-language reasoning, classification, and grounding. For referring segmentation, we use mean Intersection over Union (mIoU) on RefCOCO, RefCOCO+, RefCOCOg, and gRefCOCO, measuring the overlap between predicted masks and ground truth under language prompts. For controllable image generation and editing, we evaluate fidelity using FID (Fréchet Inception Distance), and alignment using CLIP-based scores (CLIP-I, CLIP-T, CLIP-DIR) to assess visual consistency and instruction compliance. Additional breakdowns from GenAI-bench and GenEval are used for fine-grained control metrics such as object count, color, and spatial placement.

\section{Unified Segmentation Capability of FOCUS}

FOCUS is designed as a unified model capable of handling diverse segmentation tasks spanning spatial, temporal, and interactive modalities. We evaluate its generalization and adaptability across five key segmentation scenarios, each highlighting a distinct dimension of its fine-grained visual perception.

\subsection{Contextual Reasoning and Referential Understanding}

We evaluate the ability of FOCUS to perform segmentation under semantically complex and referential conditions using the ReasonSeg and ReVOS benchmarks. These tasks require the model to accurately localize and segment objects based on contextual or linguistic descriptions with varying temporal spans. As shown in Table~\ref{tab:refseg}, FOCUS achieves leading performance across both benchmarks. On ReasonSeg, it obtains a gIoU of 62.1 and a cIoU of 58.6, surpassing prior methods in semantic segmentation precision. On ReVOS, FOCUS ranks first in all reasoning and referring metrics, including a J\&F of 57.2 in reasoning and 58.9 in referring, validating its unified capability in both spatial understanding and temporal reference tracking.


\begin{table}[ht]
\centering
\caption{Performance on Referring Video Object Segmentation (ReVOS) and ReasonSeg benchmarks. Bold: best; Underlined: second-best.}
\scalebox{0.8}{
\begin{tabular}{l|c|ccc|ccc|ccc|cc}
\hline
                         &                                                                 & \multicolumn{3}{l|}{ReVOS-Reasoning} & \multicolumn{3}{l|}{ReVOS-Referring} & \multicolumn{3}{l|}{ReVOS-Overall} & \multicolumn{2}{l}{ReasonSeg} \\ \cline{3-13} 
\multirow{-2}{*}{Method} & \multirow{-2}{*}{Backbone}                                      & J          & F          & J\&F        & J          & F          & J\&F        & J          & F         & J\&F       & gIoU          & cIoU          \\ \hline
LMPM                     & Swin-T                                                          & 13.3       & 24.3       & 18.8       & 29.0       & 39.1       & 34.1       & 21.2       & 31.7      & 26.4      & -             & -             \\
ReferFormer              & Video-Swin-B                                                    & 21.3       & 25.6       & 23.4       & 31.2       & 34.3       & 32.7       & 26.2       & 29.9      & 28.1      & -             & -             \\
LISA-7B                  & ViT-H                                                           & 33.8       & 38.4       & 36.1       & 44.3       & 47.1       & 45.7       & 39.1       & 42.7      & 40.9      & 52.9          & 54.0          \\
LaSagnA-7B               & ViT-H                                                           & -          & -          & -          & -          & -          & -          & -          & -         & -         & 48.8          & 47.2          \\
SAM4MLLM-7B              & \begin{tabular}[c]{@{}c@{}}Efficient\\ ViT-SAM-XL1\end{tabular} & -          & -          & -          & -          & -          & -          & -          & -         & -         & 46.7          & 48.1          \\
TrackGPT-13B             & ViT-H                                                           & 38.1       & 42.9       & 40.5       & 48.3       & 50.6       & 49.5       & 43.2       & 46.8      & 45.0      & -             & -             \\
VISA-7B                  & ViT-H                                                           & 36.7       & 41.7       & 39.2       & 51.1       & 54.7       & 52.9       & 43.9       & 48.2      & 46.1      & 52.7          & \textbf{57.8} \\
VISA-13B                 & ViT-H                                                           & \underline{38.3}       & \underline{43.5}       & \underline{40.9}       & \underline{52.3}       & \underline{55.8}       & \underline{54.1}       & \underline{45.3}       & \underline{49.7}      & \underline{47.5}      & -             & -             \\
HyperSeg-3B              & Swin-B                                                          & 50.2       & 55.8       & 53.0       & 56.0       & 60.9       & 58.5       & 53.1       & 58.4      & 55.7      & \underline{59.2}          & 56.7          \\ \hline
\rowcolor[HTML]{CBCEFB} 
FOCUS                    & ConvNext-L                                                      & \textbf{51.6}       & \textbf{56.3}       & \textbf{57.2}       & \textbf{56.8}       & \textbf{61.0}       & \textbf{58.9}       & \textbf{54.3}       & \textbf{59.1}      & \textbf{56.7}      & \textbf{62.1}          & \underline{58.6}          \\ \hline
\end{tabular}}
\label{tab:refseg}
\end{table}

\subsection{User-Guided Interactive Perception}

To assess FOCUS's adaptability to user-driven segmentation, we evaluate it on the COCO-Interactive benchmark under four input modes: box, point, mask, and scribble. These interactions simulate real-time editing scenarios where users specify partial object regions.

As summarized in Table~\ref{tab:interactive}, FOCUS achieves the highest IoU across all modalities, including a box score of 83.4 and a point score of 78.6, demonstrating robust generalization across sparse and dense cues. This consistent performance under varied interaction types highlights FOCUS’s potential for responsive and controllable editing applications.


\begin{table}[ht]
\centering
\caption{Performance on the COCO-Interactive benchmark. IoU (\%) under different input modalities.}
\begin{tabular}{l|l|c|c|c|c}
\hline
Method   & Backbone   & Box & Scribble & Mask & Point \\ \hline
SAM      & ViT-B      & 68.7 & -        & -    & 33.6  \\
SAM      & ViT-L      & 71.6 & -        & -    & 37.7  \\
SEEM     & DaViT-B    & 42.1 & 44.0     & 65.0 & 57.8  \\
PSALM    & Swin-B     & \underline{80.9} & \underline{80.0}     & \underline{82.4} & \underline{74.0}  \\
HyperSeg & Swin-B     & 77.3 & 75.2     & 79.5 & 63.4  \\
\rowcolor[HTML]{CBCEFB} 
FOCUS    & ConvNext-L & \textbf{83.4} & \textbf{82.7} & \textbf{85.2} & \textbf{78.6}  \\ \hline
\end{tabular}
\label{tab:interactive}
\end{table}

\subsection{Temporal Consistency in Dynamic Scenes}
To evaluate FOCUS’s capability in video-level segmentation, we benchmark it on both \textbf{Video Object Segmentation (VOS)} and \textbf{Video Instance Segmentation (VIS)} tasks. These tasks test the model's ability to consistently track and segment objects across time, even in the presence of occlusions, motion blur, or viewpoint shifts.

Table~\ref{tab:temporal} shows that FOCUS achieves top performance across all datasets, including a J\&F score of 79.1 on DAVIS17 and a mAP of 65.7 on YouTube-VIS. Its strong performance on both generic and referring-based video benchmarks confirms its robustness in temporally consistent, instance-aware segmentation.

\begin{table}[ht]
\centering
\caption{Video segmentation results across DAVIS17 (VOS), Ref-YT, Ref-DAVIS (R-VOS), and YouTube-VIS (VIS). Bold: best; Underlined: second-best.}
\scalebox{0.85}{
\begin{tabular}{l|l|c|c|c|c}
\hline
Method                  & Backbone     & DAVIS17 (J\&F) & Ref-YT (J\&F) & Ref-DAVIS (J\&F) & YT-VIS (mAP) \\ \hline
SEEM                    & DaViT-B      & 62.8           & -            & -               & -            \\
OMG-Seg                 & ConvNeXt-L   & 74.3           & -            & -               & 56.4         \\
ReferFormer             & Video-Swin-B & -              & 62.9         & 61.1            & -            \\
OnlineRefer             & Swin-L       & -              & 63.5         & 64.8            & -            \\
UNINEXT                 & ConvNeXt-L   & 77.2           & 66.2         & 66.7            & \underline{64.3}         \\ \hline
LISA-7B                 & ViT-H        & -              & 53.9         & 64.8            & -            \\
VISA-13B                & ViT-H        & -              & 63.0         & 70.4            & -            \\
VideoLISA-3.8B          & ViT-H        & -              & 63.7         & 68.8            & -            \\
HyperSeg-3B             & Swin-B       & \underline{77.6}           & \underline{68.5}         & \underline{71.2}            & 53.8         \\ \hline
\rowcolor[HTML]{CBCEFB} 
FOCUS                   & ConvNeXt-L   & \textbf{79.1}  & \textbf{69.3} & \textbf{72.4}   & \textbf{65.7}     \\ \hline
\end{tabular}}
\label{tab:temporal}
\end{table}

\section{Visualization of Generation Quality}
\begin{figure}[ht]
    \centering
    \includegraphics[width=1\textwidth]{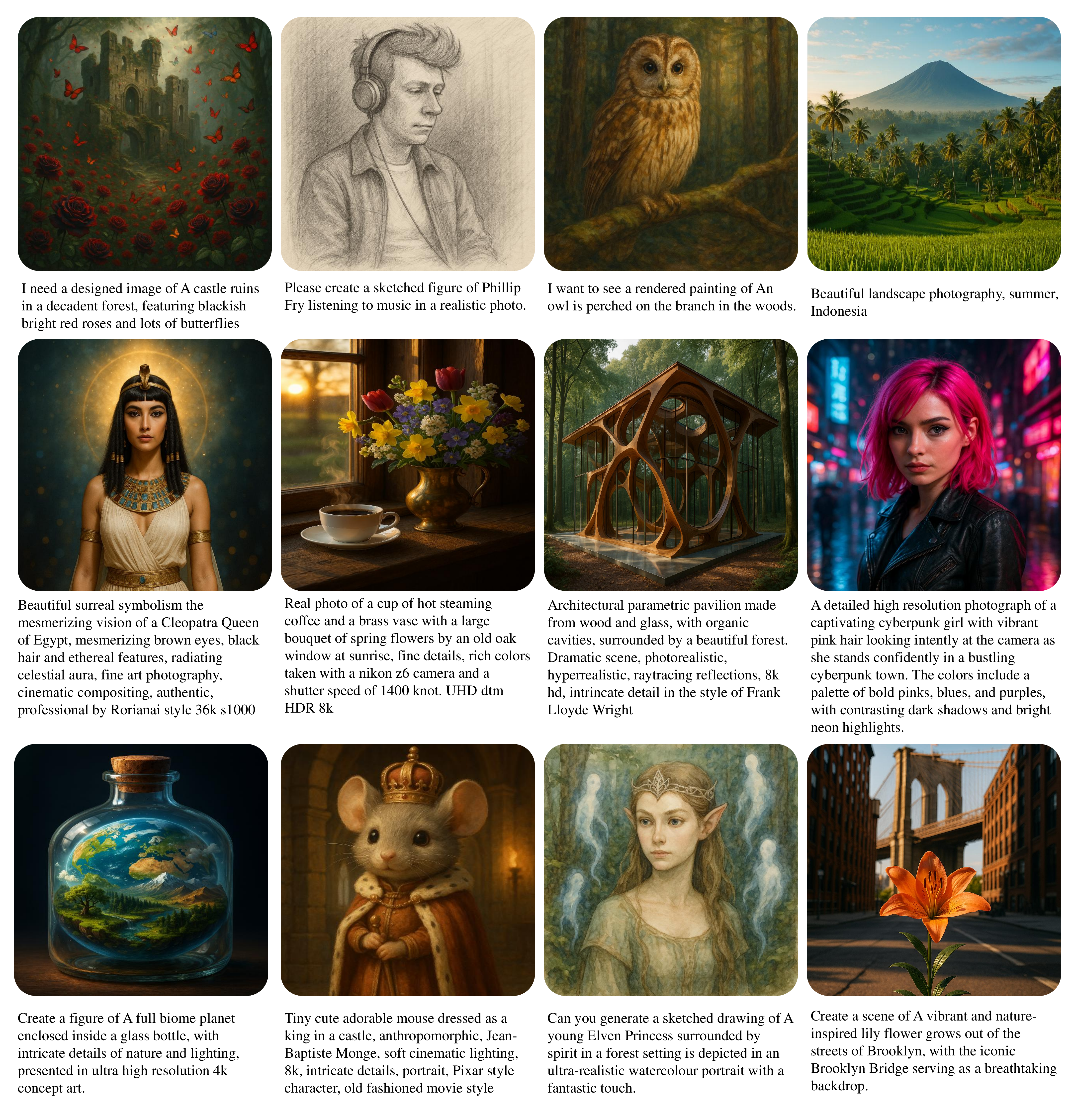}
    \caption{
Visual examples of image generation results from FOCUS given only natural language prompts. The model produces diverse and high-fidelity outputs across a range of styles and scenes, demonstrating strong semantic alignment and visual controllability.
    }
    \label{fig:visual_generation}
\end{figure}

We present a set of images generated by FOCUS based on natural language prompts (see figure \ref{fig:visual_generation}). These results span various styles including realistic scenes, conceptual designs, and artistic illustrations. They demonstrate the model's ability to produce high-quality and semantically consistent outputs.

Each image is generated solely from text input without using any visual masks or interactive guidance. The visualizations confirm FOCUS's strength in aligning language instructions with fine-grained visual content, showcasing both fidelity and controllability.

\section{Controllable Image Editing via Segmentation Guidance}

\begin{figure}[ht]
    \centering
    \includegraphics[width=1\textwidth]{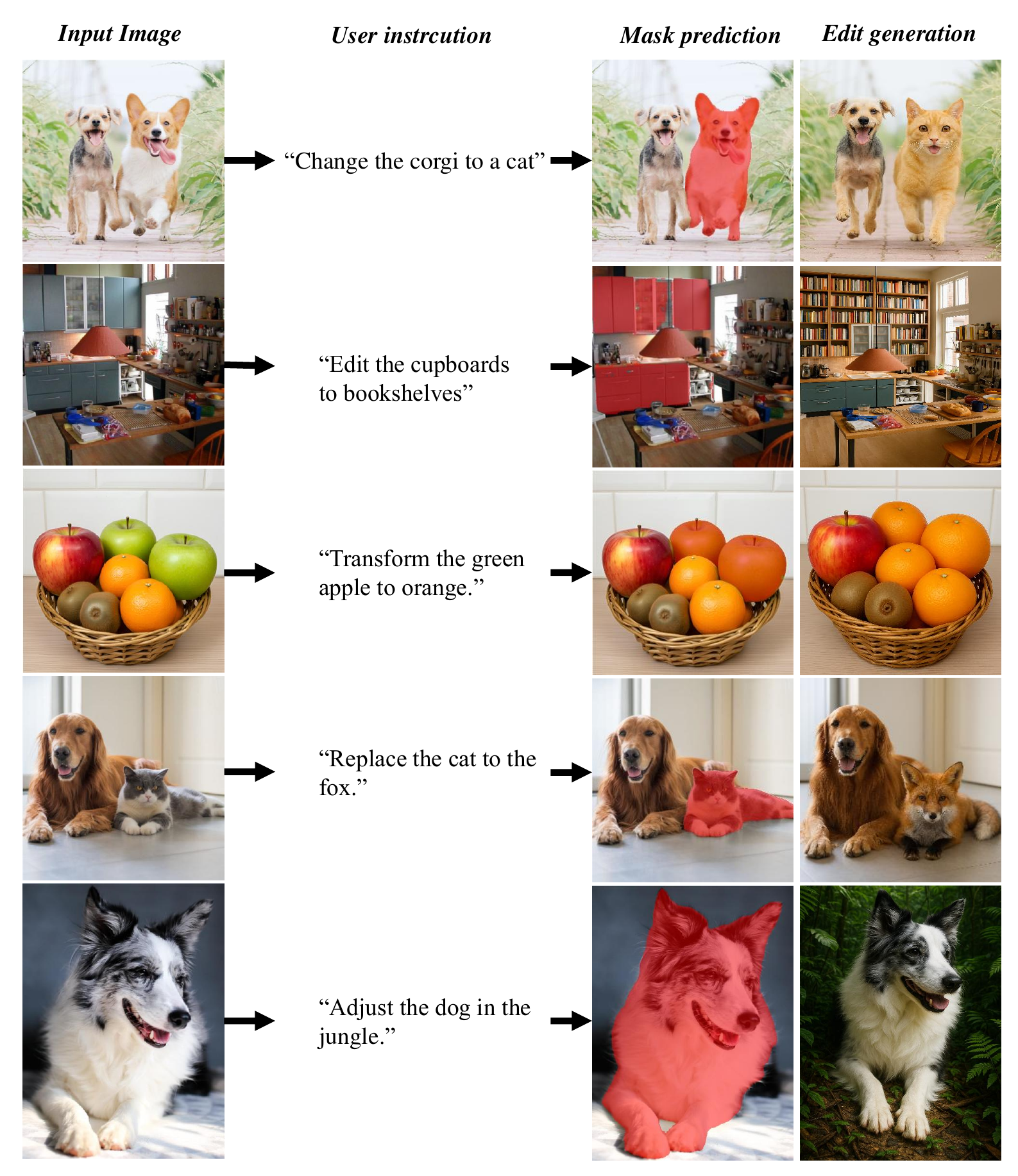}
    \caption{
Visualization of controllable image editing results. Given an input image and a user instruction, FOCUS first predicts a spatial mask corresponding to the referential target, then performs localized generation to edit the specified region. The examples demonstrate accurate region identification and high-fidelity edits aligned with the instruction.
    }
    \label{fig:visual_video}
\end{figure}

To further demonstrate the controllability of FOCUS in localized image editing, we visualize representative examples where the model edits specific regions based on natural language instructions. As shown in Figure ~\ref{fig:visual_video}, the model takes an input image and a user-provided instruction describing the desired modification. It first predicts a segmentation mask that identifies the target region referenced in the text, then uses this mask to guide the generation of an edited output.

This pipeline illustrates the effectiveness of FOCUS in grounding language to spatial regions and applying precise, content-aware modifications. The examples cover diverse scenarios including object transformation, replacement, and contextual adjustment. The results validate that segmentation-aware diffusion guidance enables fine-grained, localized editing while preserving the global scene structure.

\begin{table}[ht]
\centering
\caption{Prompt schema design for different vision-language tasks. Each row shows how FOCUS handles a specific task by pairing a general instruction (task prompt) with a task-specific condition. This formulation supports unified handling of segmentation, generation, and editing tasks.}
\scalebox{0.78}{
\begin{tabular}{l|p{5.2cm}|p{6.5cm}}
\hline
\textbf{Task Type} & \textbf{Task Instruction Prompt (SI)} & \textbf{Condition Prompt (SC)} \\
\hline
Class-based Segmentation & Please segment all the positive objects by the following candidate categories. & \texttt{["person", "dog", "car", "tree", ...]} \\
\hline
Referring Segmentation & Please segment the target referred to by the language description. & \texttt{"The man wearing a red hat standing beside the yellow car."} \\
\hline
Reasoning Segmentation & Please segment the target referred to by the reasoning-based description. & \texttt{"The object that the man is reaching for in the office."} \\
\hline
Interactive Segmentation & Please segment according to the given visual reference regions. & Pooled CLIP region features (e.g., clicks, scribbles, boxes) \\
\hline
Image Generation & Please generate an image according to the following description. & \texttt{"A tiny brown dog with white patches, eagerly holding a blue and black Frisbee."} \\
\hline
Image Editing & Please edit the image according to the following instruction. & \texttt{"Replace the man in a black jacket with a woman in the same pose."} \\
\hline
\end{tabular}
}
\label{tab:prompt}
\end{table}

\section{Prompt Design for Multi-Task Vision-Language Modeling}
To support a wide range of vision-language tasks within a unified framework, FOCUS adopts a structured prompt schema consisting of two components: a task instruction prompt and a condition prompt in Table.~\ref{tab:prompt}. The task instruction prompt defines the model objective in natural language, such as segmentation, generation, or editing. The condition prompt provides task-specific contextual information, such as category labels, referential descriptions, or visual cues.

This design enables the model to flexibly adapt to diverse tasks including class-based segmentation, referring and reasoning segmentation, interactive segmentation with visual cues, text-to-image generation, and fine-grained image editing. By standardizing task formulation through prompt schema, FOCUS achieves better generalization across modalities and applications.

\clearpage

\end{document}